\documentclass[10pt,twocolumn,letterpaper]{article}

\usepackage{cvpr}
\usepackage{times}
\usepackage{epsfig}
\usepackage{graphicx}
\usepackage{amsmath}
\usepackage{amssymb}
\usepackage{booktabs}
\usepackage[caption=false]{subfig}
\usepackage{mynotation}

\usepackage[pagebackref=true,breaklinks=true,letterpaper=true,colorlinks,bookmarks=false]{hyperref}

\newcommand{\parsection}[1]{\noindent\textbf{#1:}}

\cvprfinalcopy % *** Uncomment this line for the final submission

\def\cvprPaperID{****} % *** Enter the CVPR Paper ID here

\ifcvprfinal\fi
\begin{document}

%%%%%%%%% TITLE
\title{ECO: Efficient Convolution Operators for Tracking}

\author{Martin Danelljan, Goutam Bhat, Fahad Shahbaz Khan, Michael Felsberg \\
	\small Computer Vision Laboratory, Department of Electrical Engineering, Link\"oping University, Sweden\\
	\small\{\texttt{martin.danelljan},\; \texttt{goutam.bhat},\; \texttt{fahad.khan},\; \texttt{michael.felsberg}\}\texttt{@liu.se}
	{}
}

\maketitle
%\thispagestyle{empty}

%%%%%%%%% ABSTRACT
\begin{abstract}
   In recent years, Discriminative Correlation Filter (DCF) based methods have significantly advanced the state-of-the-art in tracking. However, in the pursuit of ever increasing tracking performance, their characteristic speed and real-time capability have gradually faded. Further, the increasingly complex models, with massive number of trainable parameters, have introduced the risk of severe over-fitting. In this work, we tackle the key causes behind the problems of computational complexity \emph{and} over-fitting, with the aim of simultaneously improving \emph{both} speed and performance.

We revisit the core DCF formulation and introduce: (i) a factorized convolution operator, which drastically reduces the number of parameters in the model; (ii) a compact generative model of the training sample distribution, that significantly reduces memory and time complexity, while providing better diversity of samples; (iii) a conservative model update strategy with improved robustness and reduced complexity. We perform comprehensive experiments on four benchmarks: VOT2016, UAV123, OTB-2015, and TempleColor.
When using expensive deep features, our tracker provides a 20-fold speedup and achieves a $13.0 \%$ relative gain in Expected Average Overlap compared to the top ranked method \cite{DanelljanECCV2016} in the VOT2016 challenge. Moreover, our fast variant, using hand-crafted features, operates at 60 Hz on a single CPU, while obtaining $65.0 \%$ AUC on OTB-2015.
\end{abstract}

%%%%%%%%% BODY TEXT

\section{Introduction}

Generic visual tracking is one of the fundamental problems in computer vision. It is the task of estimating the trajectory of a target in an image sequence, given only its initial state. Online visual tracking plays a crucial role in numerous real-time vision applications, such as smart surveillance systems, autonomous driving, UAV monitoring, intelligent traffic control, and human-computer-interfaces. Due to the online nature of tracking, an ideal tracker should be accurate and robust under the hard computational constraints of real-time vision systems.

\begin{figure}[!t]
	\centering%
	\newcommand{\wid}{0.25\columnwidth}%
	\includegraphics*[trim = 80 0 90 0, width = \wid]{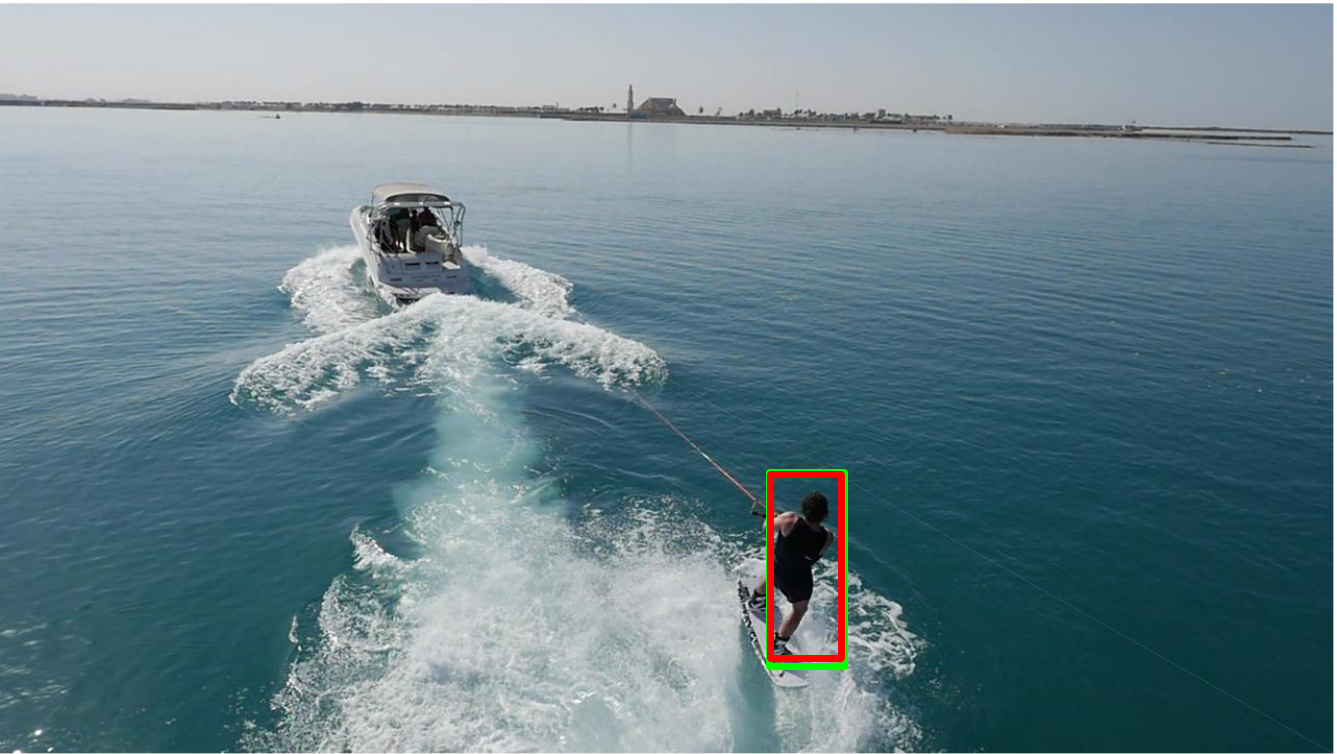}%
	\includegraphics*[trim = 90 0 80 0, width = \wid]{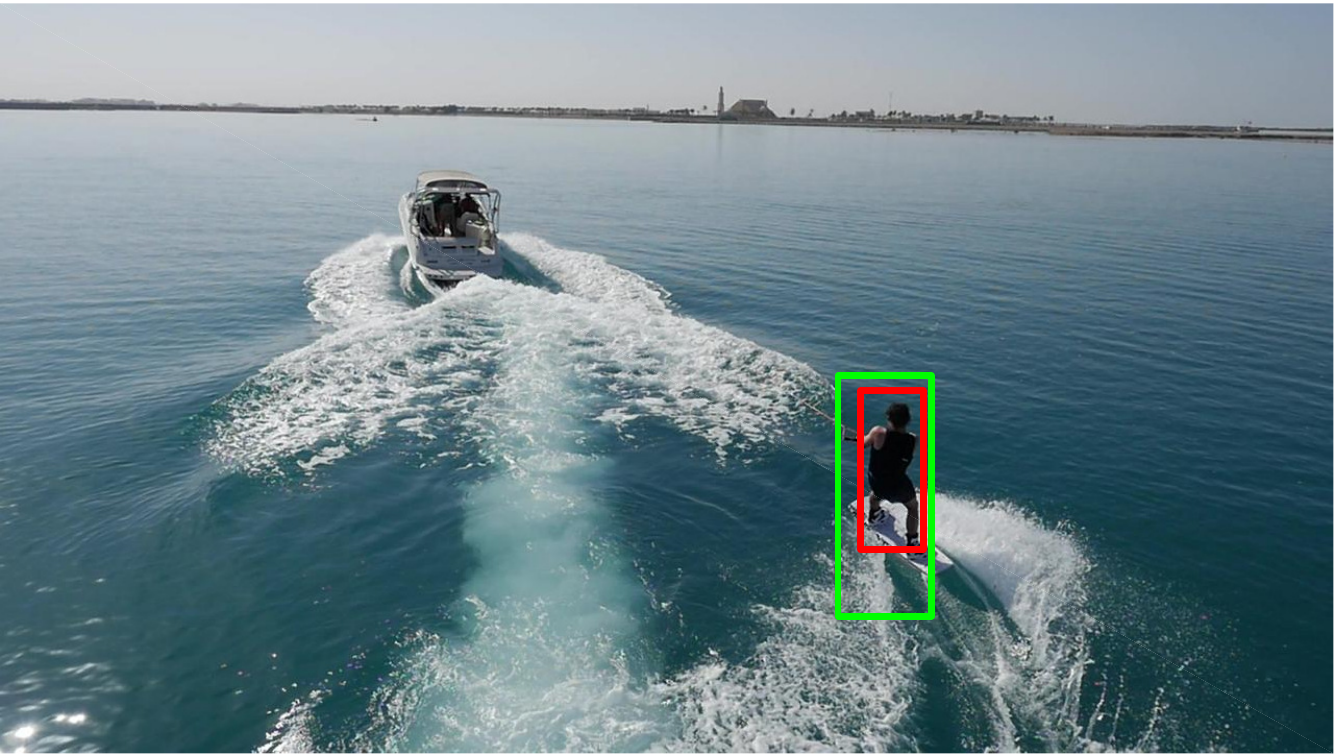}%
	\includegraphics*[trim = 80 0 90 0, width = \wid]{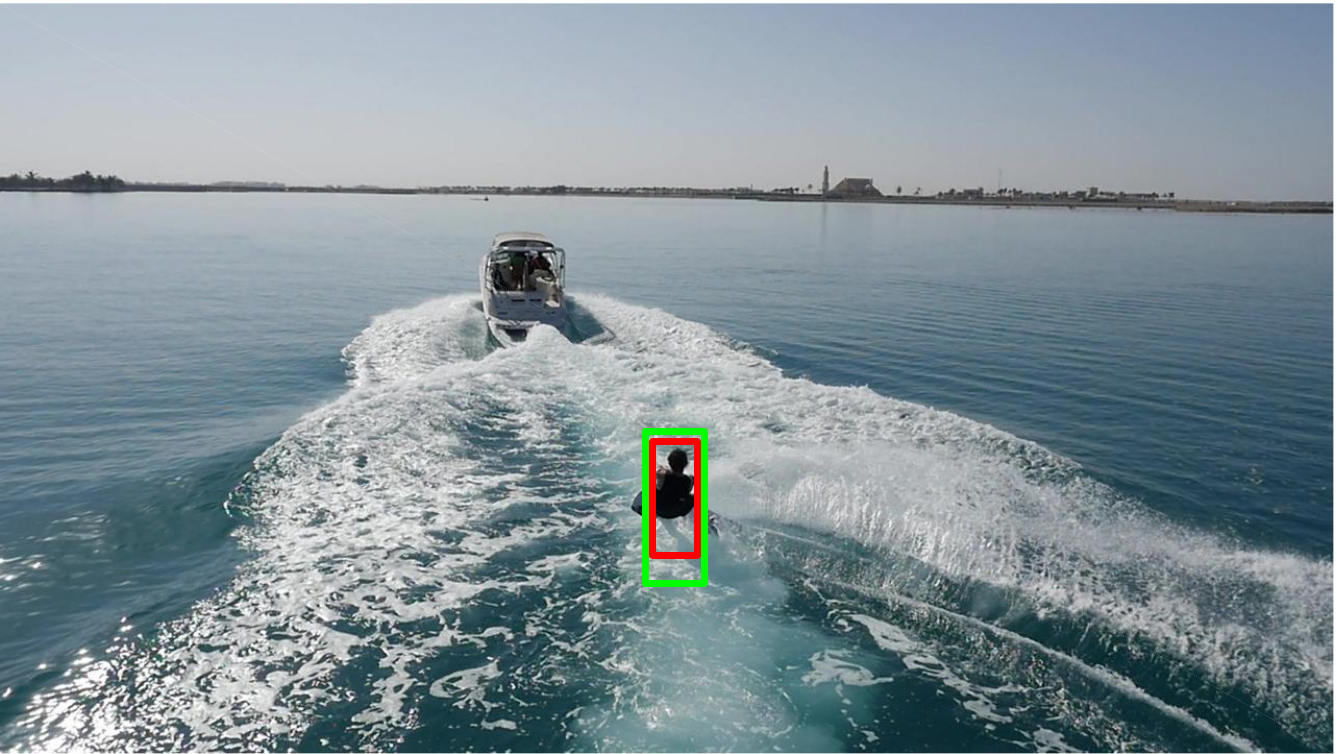}%
	\includegraphics*[trim = 30 0 140 0, width = \wid]{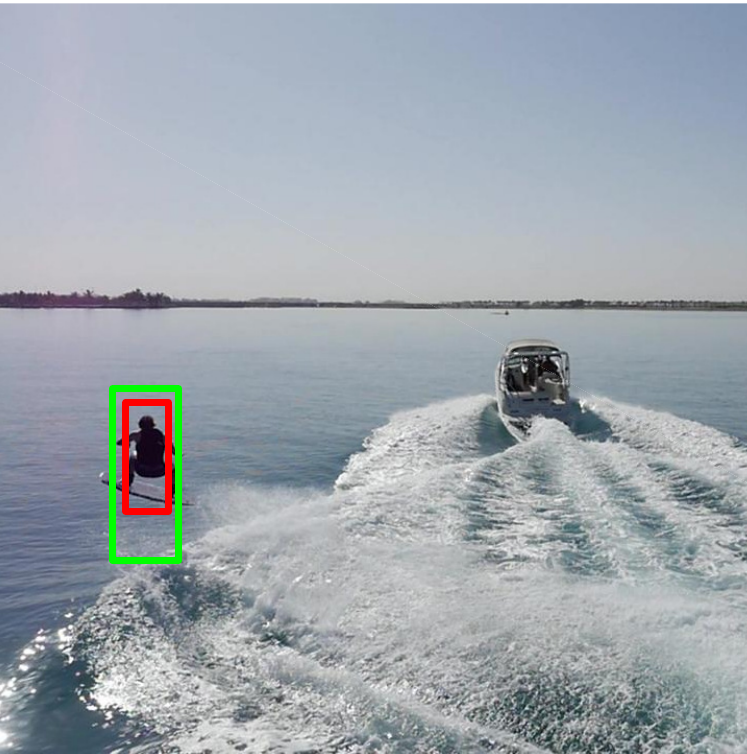}
	\includegraphics*[trim = 0 0 0 50, width = \wid]{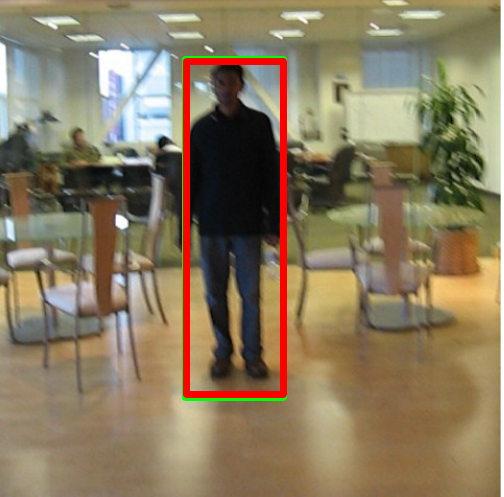}%
	\includegraphics*[trim = 0 0 0 50, width = \wid]{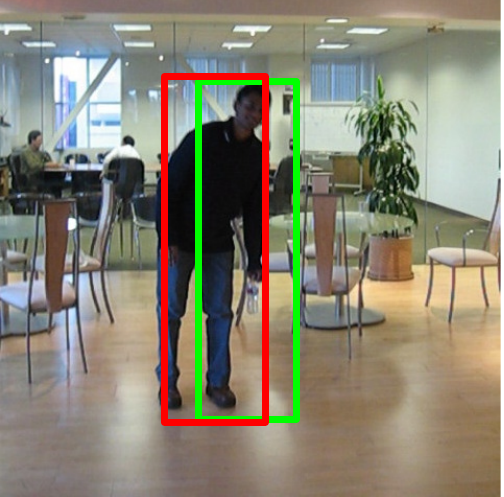}%
	\includegraphics*[trim = 0 0 0 50, width = \wid]{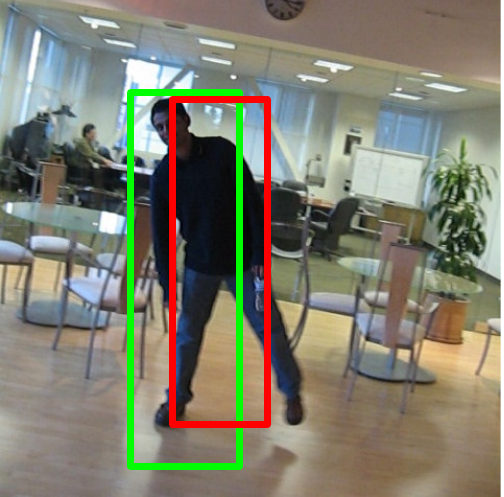}%
	\includegraphics*[trim = 0 0 0 50, width = \wid]{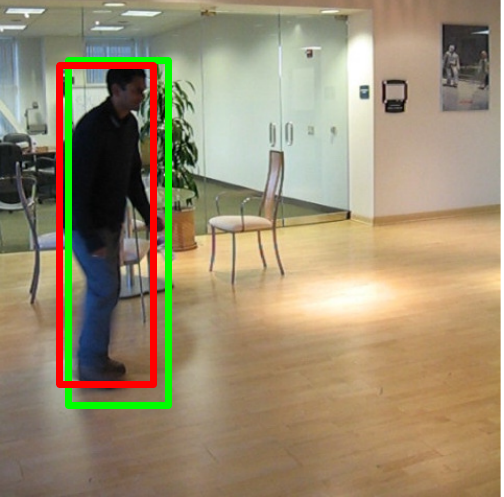}
	\includegraphics*[trim = 10 0 60 0, width = \wid]{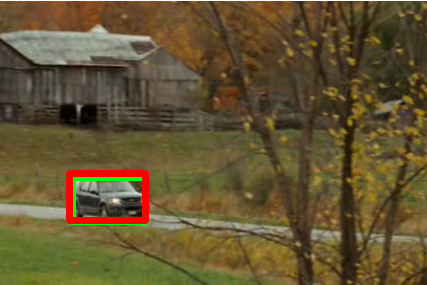}%
	\includegraphics*[trim = 10 0 60 0, width = \wid]{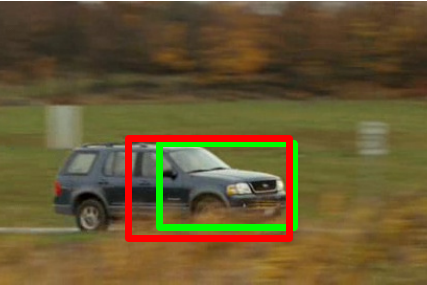}%
	\includegraphics*[trim = 10 0 60 0, width = \wid]{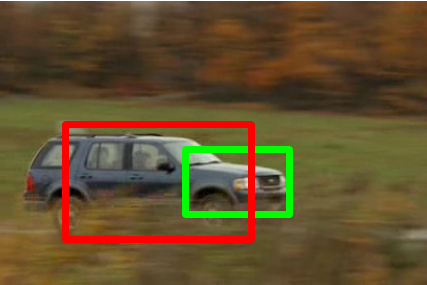}%
	\includegraphics*[trim = 10 0 60 0, width = \wid]{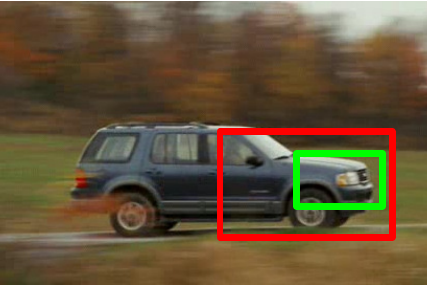}
	\includegraphics*[trim = 2 2 2 5, width = 0.42\columnwidth]{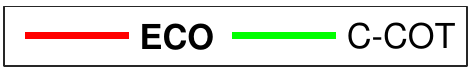}\vspace{-1mm}%
	\caption{A comparison of our approach ECO with the baseline C-COT \cite{DanelljanECCV2016} on three example sequences. In all three cases, C-COT suffers from severe over-fitting to particular regions of the target. This causes poor target estimation in cases of scale variations (top row), deformations (middle row), and out-of-plane rotations (bottom row). Our ECO tracker successfully tackles the causes of over-fitting, leading to better generalization of the target appearance, while achieving a 20-fold speedup.}\vspace{-2.5mm}%
	\label{fig:intro}%
\end{figure}

In recent years, Discriminative Correlation Filter (DCF) based approaches have shown continuous performance improvements in terms of accuracy and robustness on tracking benchmarks \cite{VOT2016,OTB2015}. The recent advancement in DCF based tracking performance is driven by the use of multi-dimensional features \cite{DanelljanCVPR14,galoogahiICCV13}, robust scale estimation \cite{DanelljanBMVC14,DanelljanTPAMI2016}, non-linear kernels \cite{Henriques14}, long-term memory components \cite{LTC_CVPR15}, sophisticated learning models \cite{KAUST_ECCV16,DanelljanCVPR2016a} and reducing boundary effects \cite{DanelljanICCV2015,GaloogahiCVPR2015}. However, these improvements in accuracy come at the price of significant reductions in tracking speed. For instance, the pioneering MOSSE tracker by Bolme \etal \cite{MOSSE2010} is about $1000 \times$ faster than the recent top-ranked DCF tracker, C-COT \cite{DanelljanECCV2016}, in the VOT2016 challenge \cite{VOT2016}, but obtains only half the accuracy.

As mentioned above, the advancement in DCF tracking performance is predominantly attributed to powerful features and sophisticated learning formulations \cite{DanelljanVOT2015,DanelljanECCV2016,HCF_ICCV15}. This has led to substantially larger models, requiring hundreds of thousands of trainable parameters. On the other hand, such complex and large models have introduced the risk of severe over-fitting (see figure~\ref{fig:intro}). In this paper, we tackle the issues of over-fitting in recent DCF trackers, while restoring their hallmark real-time capabilities.

\subsection{Motivation}

We identify three key factors that contribute to \emph{both} increased computational complexity and over-fitting in state-of-the-art DCF trackers.

\parsection{Model size}
The integration of high-dimensional feature maps, such as deep features, has led to a radical increase in the number of appearance model parameters, often beyond the dimensionality of the input image. As an example, C-COT \cite{DanelljanECCV2016} continuously updates about 800,000 parameters during the online learning of the model. Due to the inherent scarcity of training data in tracking, such a high-dimensional parameter space is prone to over-fitting. Further, the high dimensionality causes an increase in the computational complexity, leading to slow tracking speed.

\parsection{Training set size}
State-of-the-art DCF trackers, including C-COT, require a large training sample set to be stored due to their reliance on iterative optimization algorithms. In practice however, the memory size is limited, particularly when using high-dimensional features. A typical strategy for maintaining a feasible memory consumption is to discard the oldest samples. This may however cause over-fitting to recent appearance changes, leading to model drift (see figure~\ref{fig:intro}). Moreover, a large training set increases the computational burden.

\parsection{Model update}
Most DCF-based trackers apply a continuous learning strategy, where the model is updated rigorously in every frame. On the contrary, recent works have shown impressive performance without any model update, using Siamese networks \cite{SiameseFC}. Motivated by these findings, we argue that the continuous model update in state-of-the-art DCF is excessive and sensitive to sudden changes caused by, \eg, scale variations, deformations, and out-of-plane rotations (see figure~\ref{fig:intro}). This excessive update strategy causes both lower frame-rates and degradation of robustness due to over-fitting to the recent frames.

\subsection{Contributions}
We propose a novel formulation that addresses the previously listed issues of state-of-the-art DCF trackers. As our first contribution, we introduce a factorized convolution operator that dramatically reduces the number of parameters in the DCF model. Our second contribution is a compact generative model of the training sample space that effectively reduces the number of samples in the learning, while maintaining their diversity. As our final contribution, we introduce an efficient model update strategy, that simultaneously improves tracking speed and robustness. 

Comprehensive experiments clearly demonstrate that our approach concurrently improves both tracking performance and speed, thereby setting a new state-of-the-art on four benchmarks: VOT2016, UAV123, OTB-2015, and TempleColor. Our approach significantly reduces the number of model parameters by $80 \%$, training samples by $90 \%$ and optimization iterations by $80 \%$ in the learning, compared to the baseline. On VOT2016, our approach outperforms the top ranked tracker, C-COT \cite{DanelljanECCV2016}, in the challenge, while achieving a significantly higher frame-rate. Furthermore, we propose a fast variant of our tracker that maintains competitive performance, with a speed of 60 frames per second (FPS) on a single CPU, thereby being especially suitable for computationally restricted robotics platforms.

\section{Baseline Approach: C-COT}
\label{sec:CCOT}

In this work, we collectively address the problems of computational complexity and over-fitting in state-of-the-art DCF trackers. We adopt the recently introduced Continuous Convolution Operator Tracker (C-COT) \cite{DanelljanECCV2016} as our baseline. The C-COT obtained the top rank in the recent VOT2016 challenge \cite{VOT2016}, and has demonstrated outstanding results on other tracking benchmarks \cite{TempleColor,OTB2015}. Unlike the standard DCF formulation, Danelljan \etal \cite{DanelljanECCV2016} pose the problem of learning the filters in the continuous spatial domain. The generalized formulation in C-COT yields two advantages that are relevant to our work.

The first advantage of C-COT is the natural integration of multi-resolution feature maps, achieved by performing convolutions in the continuous domain. This provides the flexibility of choosing the cell size (\ie resolution) of each visual feature independently, without the need for explicit re-sampling.
The second advantage is that the predicted detection scores of the target are directly obtained as a continuous function, enabling accurate sub-grid localization.

Here, we briefly describe the C-COT formulation, adopting the same notation as in \cite{DanelljanECCV2016} for convenience. The C-COT discriminatively learns a convolution filter based on a collection of $M$ training samples $\{x_j\}_1^M \subset \mathcal{X}$. Unlike the standard DCF, each feature layer $x_j^d \in \reals^{N_d}$ has an independent resolution $N_d$.\footnote{For clarity, we present the one-dimensional domain formulation. The generalization to higher dimensions, including images, is detailed in \cite{DanelljanECCV2016}.} The feature map is transfered to the continuous spatial domain $t \in [0, T)$ by introducing an interpolation model, given by the operator $J_d$,
\begin{equation}
\label{eq:interp_op}
J_d \big\{x^d\big\} (t) = \sum_{n=0}^{N_d-1} x^d[n] b_d\left(t - \frac{T}{N_d} n\right) .
\end{equation}
Here, $b_d$ is an interpolation kernel with period $T > 0$. The result $J_d \big\{x^d\big\}$ is thus an interpolated feature layer, viewed as a continuous $T$-periodic function. We use $J \{x\}$ to denote the entire interpolated feature map, where $J \{x\}(t) \in \reals^D$.

In the C-COT formulation, a continuous $T$-periodic multi-channel convolution filter $f = (f^1 \ldots f^D)$ is trained to predict the detection scores $S_f\{x\}(t)$ of the target as,
\begin{equation}
\label{eq:conv_op}
S_f\{x\} = f \conv J \{x\} = \sum_{d=1}^D f^d \conv J_d \big\{x^d\big\} \,.
\end{equation}
The scores are defined in the corresponding image region $t \in [0, T)$ of the feature map $x \in \mathcal{X}$. In \eqref{eq:conv_op}, the convolution of single-channel $T$-periodic functions is defined as $f \conv g (t) = \frac{1}{T} \integlim{0}{T} f(t - \tau) g(\tau) \diff \tau$. The multi-channel convolution $f \conv J \{x\}$ is obtained by summing the result of all channels, as defined in \eqref{eq:conv_op}. The filters are learned by minimizing the following objective,
\begin{equation}
\label{eq:loss_spatial}
E(f) = \sum_{j=1}^{M} \alpha_j \left\| S_f\{x_j\} - y_j \right\|^2_{L^2} + \sum_{d=1}^{D} \left\| w f^d \right\|^2_{L^2} \,.
\end{equation}
The labeled detection scores $y_j(t)$ of sample $x_j$ is set to a periodically repeated Gaussian function. The data term consists of the weighted classification error, given by the $L^2$-norm $\|g\|^2_{L^2} = \frac{1}{T} \integlim{0}{T} |g(t)|^2 \diff t$, where $\alpha_j \geq 0$ is the weight of sample $x_j$. The regularization integrates a spatial penalty $w(t)$ to mitigate the drawbacks of the periodic assumption, while enabling an extended spatial support \cite{DanelljanICCV2015}.

As in previous DCF methods, a more tractable optimization problem is obtained by changing to the Fourier basis. Parseval's formula implies the equivalent loss,
\begin{equation}
\label{eq:loss_fs}
E(f) = \sum_{j=1}^{M} \alpha_j \left\| \widehat{S_f\{x_j\}} - \hat{y}_j \right\|^2_{\ell^2} + \sum_{d=1}^{D} \left\|\hat{w} \conv \hat{f}^d \right\|^2_{\ell^2} \,.
\end{equation}
Here, the hat $\hat{g}$ of a $T$-periodic function $g$ denotes the Fourier series coefficients $\hat{g}[k] = \frac{1}{T} \integlim{0}{T} g(t) e^{-i \frac{2 \pi}{T} k t} \diff t$ and the $\ell^2$-norm is defined by $\| \hat{g} \|^2_{\ell^2} = \sum_{-\infty}^{\infty} |\hat{g}[k]|^2$. The Fourier coefficients of the detection scores \eqref{eq:conv_op} are given by the formula $\widehat{S_f\{x\}} = \sum_{d=1}^{D} \hat{f}^d X^d \hat{b}_d$, where $X^d$ is the Discrete Fourier Transform (DFT) of $x^d$.

In practice, the filters $f^d$ are assumed to have finitely many non-zero Fourier coefficients $\{\hat{f}^d[k]\}_{-K_d}^{K_d}$, where $K_d = \left\lfloor \frac{N_d}{2} \right\rfloor$.
Eq.~\eqref{eq:loss_fs} then becomes a quadratic problem, optimized by solving the normal equations,
\begin{equation}
\label{eq:normal_eq}
\left( A\ctp \Gamma A + W\ctp W \right) \vecft{f} = A\ctp \Gamma \vecft{y} \,.
\end{equation}
Here, $\vecft{f}$ and $\vecft{y}$ are vectorizations of the Fourier coefficients in $f^d$ and $y_j$, respectively. The matrix $A$ exhibits a sparse structure, with diagonal blocks containing elements of the form $X^d_j[k] \hat{b}_d[k]$. Further, $\Gamma$ is a diagonal matrix of the weights $\alpha_j$ and $W$ is a convolution matrix with the kernel $\hat{w}[k]$. The C-COT \cite{DanelljanECCV2016} employs the Conjugate Gradient (CG) method \cite{NumericalOptimization} to iteratively solve \eqref{eq:normal_eq}, since it was shown to effectively utilize the sparsity structure of the problem.

\section{Our Approach}
\label{sec:method}

\begin{figure*}[!t]
	\centering%
	\newcommand{\wid}{4.23cm}\vspace{-3.5mm}%
	\subfloat[C-COT\label{fig:filters_ccot}]{\includegraphics[height = \wid]{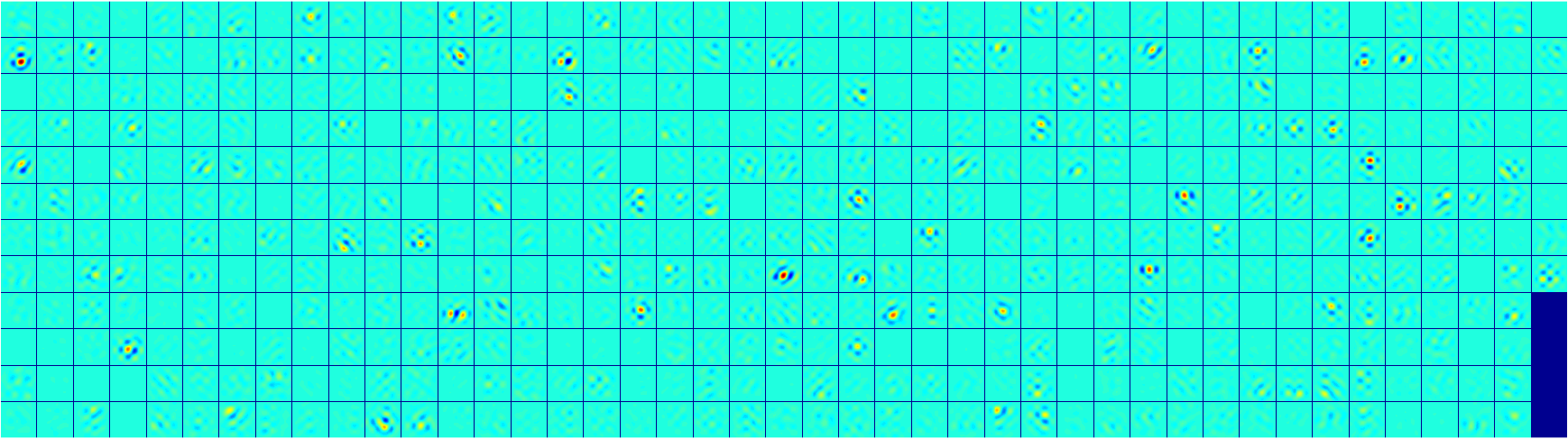}}\hspace{1mm}
	\subfloat[Ours\label{fig:filters_scot}]{\includegraphics[height = \wid]{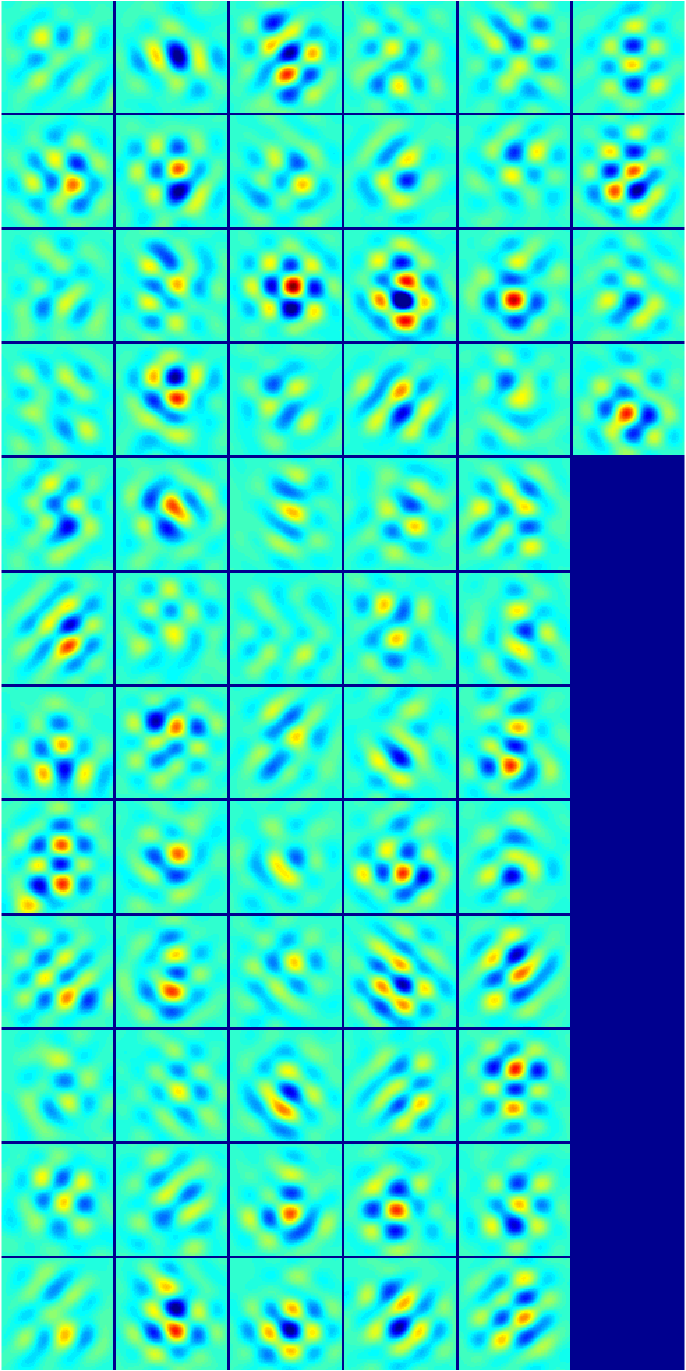}}%
	\caption{Visualization of the learned filters corresponding to the last convolutional layer in the deep network. We display all the 512 filters $f^d$ learned by the baseline C-COT (a) and the reduced set of 64 filters $f^c$ obtained by our factorized formulation (b). The vast majority of the baseline filters contain negligible energy, indicating irrelevant information in the corresponding feature layers. Our factorized convolution formulation learns a compact set of discriminative basis filters with significant energy, achieving a radical reduction of parameters.}\vspace{-2mm}%
	\label{fig:filters}
\end{figure*}

As discussed earlier, over-fitting and computational bottlenecks in the DCF learning stem from common factors. We therefore proceed with a collective treatment of these issues, aiming at both improved performance \emph{and} speed. 

\parsection{Robust learning}
As mentioned earlier, the large number of optimized parameters in \eqref{eq:loss_spatial} may cause over-fitting due to limited training data. We alleviate this issue by introducing a factorized convolution formulation in section~\ref{sec:fac}. This strategy radically reduces the number of model parameters by $80 \%$ in the case of deep features, while increasing tracking performance. Moreover, we propose a compact generative model of the sample distribution in section~\ref{sec:sample_model}, that boosts diversity and avoids the previously discussed problems related to storing a large sample set. Finally, we investigate strategies for updating the model in section~\ref{sec:optimization} and conclude that a less frequent update of the filter stabilizes the learning, which results in more robust tracking.

\parsection{Computational complexity}
The learning step is the computational bottleneck in optimization-based DCF trackers, such as C-COT. The computational complexity of the appearance model optimization in C-COT is obtained by analyzing the Conjugate Gradient algorithm applied to \eqref{eq:normal_eq}. The complexity can be expressed as $\ordo(N_\text{CG} D M \bar{K})$,\footnote{See the supplementary material for a derivation.} where $N_\text{CG}$ is the number of CG iterations and $\bar{K} = \frac{1}{D} \sum_d K_d$ is the average number of Fourier coefficients per filter channel. Motivated by this complexity analysis of the learning, we propose methods for reducing $D$, $M$ and $N_\text{CG}$ in sections \ref{sec:fac}, \ref{sec:sample_model}, and \ref{sec:optimization} respectively.

\subsection{Factorized Convolution Operator}
\label{sec:fac}

We first introduce a factorized convolution approach, with the aim of reducing the number of parameters in the model. We observed that many of the filters $f^d$ learned in C-COT contain negligible energy. This is particularly apparent for high-dimensional deep features, as visualized in figure~\ref{fig:filters}. Such filters hardly contribute to target localization, but still affect the training time. Instead of learning one separate filter for each feature channel $d$, we use a smaller set of basis filters $f^1, \ldots, f^C$, where $C < D$. The filter for feature layer $d$ is then constructed as a linear combination $\sum_{c=1}^C p_{d,c} f^c$ of the filters $f^c$ using a set of learned coefficients $p_{d,c}$. The coefficients can be compactly represented as a $D \times C$ matrix $P = (p_{d,c})$. The new multi-channel filter can then be written as the matrix-vector product $P f$. We obtain the factorized convolution operator,
\begin{equation}
\label{eq:fac_conv_op}
S_{Pf}\{x\} = Pf \conv J \{x\} \! = \! \sum_{c,d} p_{d,c} f^c \conv J_d \big\{x^d\big\} \! = \! f \conv P\tp J \{x\}.
\end{equation}
The last equality follows from the linearity of convolution. The factorized convolution \eqref{eq:fac_conv_op} can thus alternatively be viewed as a two-step operation where the feature vector $J \{x\}(t)$ at each location $t$ is first multiplied with the matrix $P\tp$. The resulting $C$-dimensional feature map is then convolved with the filter $f$. The matrix $P\tp$ thus resembles a linear dimensionality reduction operator, as used in \eg \cite{DanelljanCVPR14}. The key difference is that we learn the filter $f$ and matrix $P$ \emph{jointly}, in a discriminative fashion, by minimizing the classification error \eqref{eq:loss_spatial} of the factorized operator \eqref{eq:fac_conv_op}. 

For simplicity, we consider learning the factorized operator \eqref{eq:fac_conv_op} from single training sample $x$. To simplify notation, we use $\hat{z}^d[k] = X^d[k] \hat{b}_d[k]$ to denote the Fourier coefficients of the interpolated feature map $z = J\{x\}$. The corresponding loss in the Fourier domain \eqref{eq:loss_fs} is derived as,
\begin{equation}
	\label{eq:loss_fac}
	E(f,P) = \left\|\hat{z}\tp P \hat{f} - \hat{y}\right\|^2_{\ell^2} \! + \sum_{c=1}^{C} \left\|\hat{w} \conv \hat{f}^c \right\|^2_{\ell^2} \! + \lambda \|P\|_F^2 .
\end{equation}
Here we have added the Frobenius norm of $P$ as a regularization, controlled by the weight parameter $\lambda$.

Unlike the original formulation \eqref{eq:loss_fs}, our new loss \eqref{eq:loss_fac} is a non-linear least squares problem. Due to the bi-linearity of $\hat{z}\tp P \hat{f}$, the loss \eqref{eq:loss_fac} is similar to a matrix factorization problem \cite{Hong_2015_ICCV}. Popular optimization strategies for these applications, including Alternating Least Squares, are however not feasible due to the parameter size and online nature of our problem. Instead, we employ Gauss-Newton \cite{NumericalOptimization} and use the Conjugate Gradient method to optimize the quadratic subproblems. The Gauss-Newton method is derived by linearizing the residuals in \eqref{eq:loss_fac} using a first order Taylor series expansion. Here, this corresponds to approximating the bi-linear term $\hat{z}\tp P \hat{f}$ around the current estimate $(\hat{f}_i, P_i)$ as,
\begin{align}
\label{eq:taylor}
\hat{z}\tp (P_i + \Delta P) (\hat{f}_i + \Delta {\hat{f}}) & \approx \hat{z}\tp P_i \hat{f}_{i,\Delta} + \hat{z}\tp \Delta P \hat{f}_i \\
&= \hat{z}\tp P_i \hat{f}_{i,\Delta} + (\hat{f}_i \otimes \hat{z})\tp \vecop(\Delta P) . \nonumber
\end{align}
Here, we set $\hat{f}_{i,\Delta} = \hat{f}_i + \Delta {\hat{f}}$. In the last equality, the Kronecker product $\otimes$ is used to obtain a vectorization of the matrix step $\Delta P$.

The Gauss-Newton subproblem at iteration $i$ is derived by substituting the first-order approximation \eqref{eq:taylor} into \eqref{eq:loss_fac},
\begin{align}
	\label{eq:GN_loss}
	\tilde{E}(\hat{f}_{i,\Delta}, \Delta P) &= \left\|\hat{z}\tp P_i \hat{f}_{i,\Delta} + (\hat{f}_i \otimes \hat{z})\tp \vecop(\Delta P) - \hat{y}\right\|^2_{\ell^2} \nonumber \\ &+ \sum_{c=1}^{C} \left\|\hat{w} \conv \hat{f}_{i,\Delta}^c \right\|^2_{\ell^2} \! + \mu \|P_i + \Delta P\|_F^2 .
\end{align}
Since the filter $f$ is constrained to have finitely many non-zero Fourier coefficients, eq.\ \eqref{eq:GN_loss} is a linear least squares problem. The corresponding normal equations have a partly similar structure to \eqref{eq:normal_eq}, with additional components corresponding to the matrix increment $\Delta P$ variable.\footnote{See supplementary material for the derivation of the normal equations.} We employ the Conjugate Gradient method to optimize each Gauss-Newton subproblem to obtain the new filter $\hat{f}_{i,\Delta}^*$ and matrix increment $\Delta P^*$. The filter and matrix estimates are then updated as $\hat{f}_{i+1} = \hat{f}_{i,\Delta}^*$ and $P_{i+1} = P_i + \Delta P^*$.

The main objective of our factorized convolution operation is to reduce the computational and memory complexity of the tracker. Due to the adaptability of the filter, the matrix $P$ can be learned just from the first frame. This has two important implications. Firstly, only the projected feature map $P\tp J\{x_j\}$ requires storage, leading to significant memory savings. Secondly, the filter can be updated in subsequent frames using the projected feature maps $P\tp J\{x_j\}$ as input to the method described in section~\ref{sec:CCOT}. This reduces the linear complexity in the feature dimensionality $D$ to the filter dimensionality $C$, \ie $\ordo(N_\text{CG} C M \bar{K})$.

\begin{figure*}[!t]
	\centering%
	\newcommand{\wid}{0.86\textwidth}%
	\includegraphics[width = \wid]{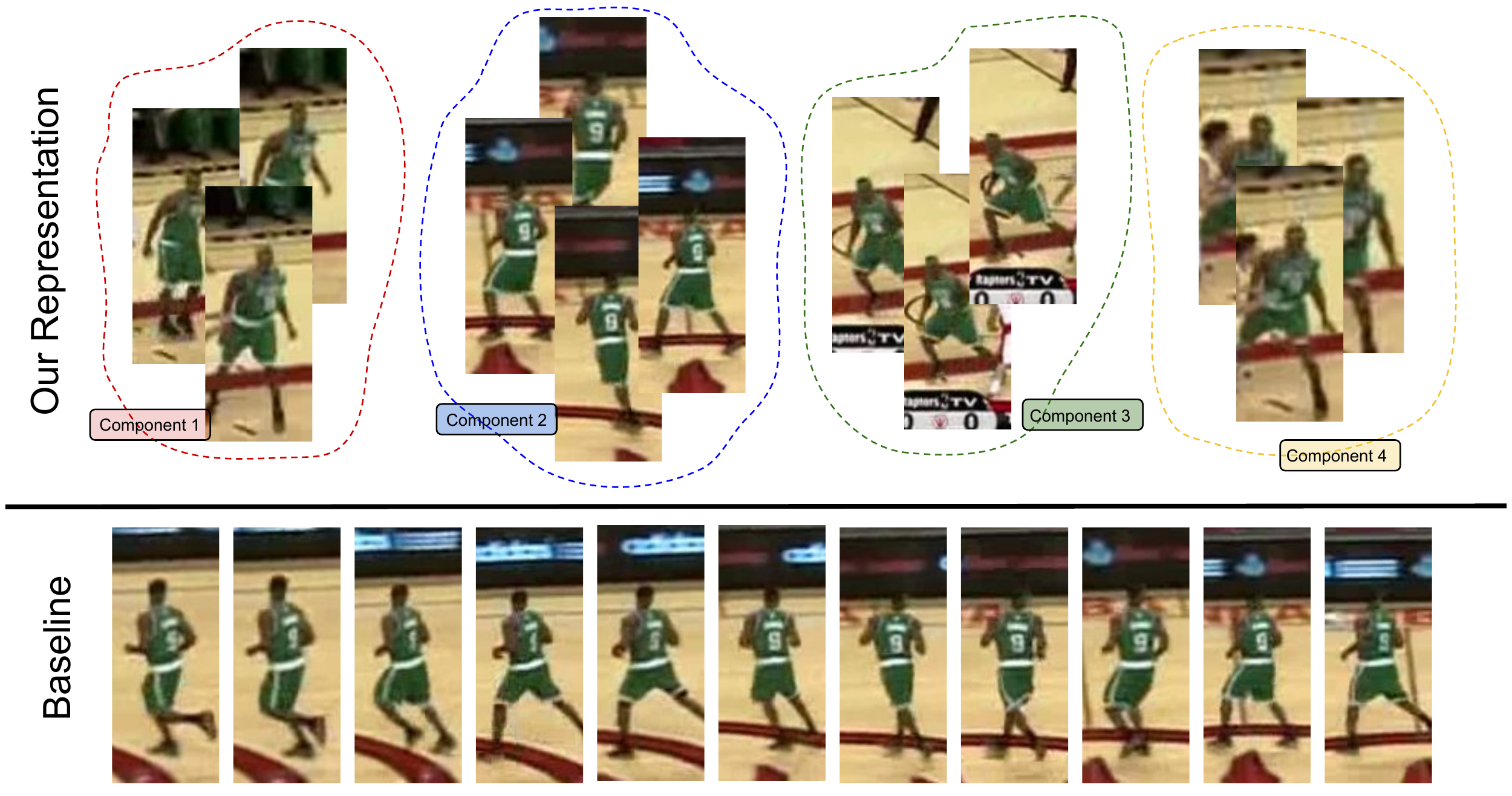}\vspace{-2mm}
	\caption{Visualization of the training set representation in the baseline C-COT (bottom row) and our method (top row). In C-COT, the training set consists of a sequence of consecutive samples. This introduces large redundancies due to slow change in appearance, while previous aspects of the appearance are forgotten. This can cause over-fitting to recent samples. Instead, we model the training data as a mixture of Gaussian components, where each component represent a different aspect of the appearance. Our approach yields a compact yet diverse representation of the data, thereby reducing the risk of over-fitting.
	}\vspace{-3mm}%
	\label{fig:sample_model}
\end{figure*}

\subsection{Generative Sample Space Model}
\label{sec:sample_model}

Here, we propose a compact generative model of the sample set that averts the earlier discussed issues of storing a large set of recent training samples. Most DCF trackers, such as SRDCF~\cite{DanelljanICCV2015} and C-COT~\cite{DanelljanECCV2016}, add one training sample $x_j$ in each frame $j$. The weights are typically set to decay exponentially $\alpha_j \sim (1 - \gamma)^{M-j}$, controlled by the learning rate $\gamma$. If the number of samples has reached a maximum limit $M_\text{max}$, the sample with the smallest weight $\alpha_j$ is replaced. This strategy however requires a large sample limit $M_\text{max}$ to obtain a representative sample set.

We observe that collecting a new sample in each frame leads to large redundancies in the sample set, as visualized in figure~\ref{fig:sample_model}. The standard sampling strategy (bottom row) populates the whole training set with similar samples $x_j$, despite containing almost the same information. Instead, we propose to use a probabilistic generative model of the sample set that achieves a compact description of the samples by eliminating redundancy and enhancing variety (top).

Our approach is based on the joint probability distribution $p(x,y)$ of the sample feature maps $x$ and corresponding desired outputs scores $y$. Given $p(x,y)$, the intuitive objective is to find the filter that minimizes the expected correlation error. This is obtained by replacing \eqref{eq:loss_spatial} with% the following loss,
\begin{equation}
\label{eq:loss_expected}
E(f) = \expec \left\{ \left\| S_f\{x\} - y \right\|^2_{L^2} \right\} + \sum_{d=1}^{D} \left\| w f^d \right\|^2_{L^2} \,.
\end{equation}
Here, the expectation $\expec$ is evaluated over the joint sample distribution $p(x,y)$. Note that the original loss \eqref{eq:loss_spatial} is obtained as a special case by estimating the sample distribution as $p(x,y) = \sum_{j=1}^{M} \alpha_j \delta_{x_j,y_j}(x,y)$, where $\delta_{x_j,y_j}$ denotes the Dirac impulse at the training sample $(x_j,y_j)$.\footnote{We can without loss of generality assume the weights $\alpha_j$ sum to one.} 
Instead, we propose to estimate a compact model of the sample distribution $p(x,y)$ that leads to a more efficient approximation of the expected loss \eqref{eq:loss_expected}.

We observe that the shape of the desired correlation output $y$ for a sample $x$ is predetermined, here as a Gaussian function. The label functions $y_j$ in \eqref{eq:loss_spatial} only differ by a translation that aligns the peak with the target center. This alignment is equivalently performed by shifting the feature map $x$. We can thus assume that the target is centered in the image region and that all $y=y_0$ are identical. 
Hence, the sample distribution can be factorized as $p(x,y) = p(x) \delta_{y_0}(y)$ and we only need to estimate $p(x)$.
For this purpose we employ a Gaussian Mixture Model (GMM) 
such that $p(x) = \sum_{l=1}^{L} \pi_l \norm(x; \mu_l; I)$. Here, $L$ is the number of Gaussian components $\norm(x; \mu_l; I)$, $\pi_l$ is the prior weight of component $l$, and $\mu_l \in \mathcal{X}$ is its mean. The covariance matrix is set to the identity matrix $I$ to avoid costly inference in the high-dimensional sample space.

To update the GMM, we use a simplified version of the online algorithm by Declercq and Piater~\cite{Declercq2008}. Given a new sample $x_j$, we first initialize a new component $m$ with $\pi_m = \gamma$ and $\mu_m = x_j$ (\emph{concatenate} in~\cite{Declercq2008}). If the number of components exceeds the limit $L$, we \emph{simplify} the GMM.  We discard a component if its weight $\pi_l$ is below a threshold. Otherwise, we merge the two closest components $k$ and $l$ into a common component $n$~\cite{Declercq2008},
\begin{equation}
	\label{eq:gmm_merge}
	\pi_n = \pi_k + \pi_l \quad , \quad \mu_n = \frac{\pi_k \mu_k + \pi_l \mu_l}{\pi_k + \pi_l} .
\end{equation}
The required distance comparisons $\|\mu_k - \mu_l\|$ are efficiently computed in the Fourier domain using Parseval's formula. Finally, the expected loss \eqref{eq:loss_expected} is approximated as,
\begin{equation}
\label{eq:loss_gmm}
E(f) = \sum_{l=1}^{L} \pi_l \left\| S_f\{\mu_l\} - y_0 \right\|^2_{L^2} + \sum_{d=1}^{D} \left\| w f^d \right\|^2_{L^2} \,.
\end{equation}
Note that the Gaussian means $\mu_l$ and the prior weights $\pi_l$ directly replace $x_j$ and $\alpha_j$, respectively, in \eqref{eq:loss_spatial}. So, the same training strategy as described in section~\ref{sec:CCOT} can be applied.

The key difference in complexity compared to \eqref{eq:loss_spatial} is that the number of samples has decreased from $M$ to $L$. In our experiments, we show that the number of components $L$ can be set to $M / 8$, while obtaining an improved tracking performance. Our sample distribution model $p(x,y)$ is combined with the factorized convolution from section~\ref{sec:fac} by replacing the sample $x$  with the projected sample $P\tp J{x}$. The projection does not affect our formulation since the matrix $P$ is constant after the first frame.

\subsection{Model Update Strategy}
\label{sec:optimization}

The standard approach in DCF based tracking is to update the model in each frame \cite{MOSSE2010,DanelljanICCV2015,Henriques14}. In C-COT, this implies optimizing \eqref{eq:loss_spatial} after each new sample is added, by iteratively solving the normal equations \eqref{eq:normal_eq}. Iterative optimization based DCF methods exploit that the loss function changes gradually between frames. The current estimate of the filter therefore provides a good initialization of the iterative search. Still, updating the filter in each frame have a severe impact on the computational load.

Instead of updating the model in a continuous fashion every frame, we use a sparser updating scheme, which is a common practice in non-DCF trackers \cite{MDNet,MEEM2014}. Intuitively, an optimization process should only be started once sufficient change in the objective has occurred. However, finding such conditions is non-trivial and may lead to unnecessarily complex heuristics. Moreover, optimality conditions based on the gradient of the loss \eqref{eq:loss_spatial}, given by the residual of \eqref{eq:normal_eq}, are expensive to evaluate in practice. We therefore avoid explicitly detecting changes in the objective and simply update the filter by starting the optimization process in every $N_\text{S}$th frame. The parameter $N_\text{S}$ determines how often the filter is updated, where $N_\text{S} = 1$ corresponds to optimizing the filter in every frame, as in standard DCF methods. In every $N_\text{S}$th frame, we perform a fixed number of $N_\text{CG}$ Conjugate Gradient iterations to refine the model. As a result, the average number of CG iterations per frame is reduced to $N_\text{CG} / N_\text{S}$, which has a substantial effect on the overall computational complexity of the learning. Note that $N_\text{S}$ does not affect the updating of the sample space model, introduced in section~\ref{sec:sample_model}, which is updated every frame. 

To our initial surprise, we observed that a moderately infrequent update of the model ($N_\text{S} \approx 5$) generally led to improved tracking results. We mainly attribute this effect to reduced over-fitting to the recent training samples. By postponing the model update a few frames, the loss is updated by adding a new mini-batch to the training samples, instead of only a single one. This might contribute to stabilizing the learning, especially in scenarios where a new sample is affected by sudden changes, such as out-of-plane rotations, deformations, clutter, and occlusions (see figure~\ref{fig:intro}).

While increasing $N_\text{S}$ leads to reduced computations, it may also reduce the convergence speed of the optimization, resulting in a less discriminative model. A naive compensation by increasing the number of CG iterations $N_\text{CG}$ would counteract the achieved computational gains. Instead, we aim to achieve a faster convergence by better adapting the CG algorithm to online tracking, where the loss changes dynamically. This is obtained by substituting the standard Fletcher-Reeves formula to the Polak-Ribi\`ere formula \cite{CGpain} for finding the momentum factor, since it has shown improved convergence rates for inexact and flexible preconditioning \cite{IPCG}, which have similarities to our scenario.

\section{Experiments}
We validate our proposed formulation by performing comprehensive experiments on
four benchmarks: VOT2016 \cite{VOT2016}, UAV123 \cite{UAV123}, OTB-2015 \cite{OTB2015}, and TempleColor \cite{TempleColor}.

\begin{table}[!t]
	\centering
	\resizebox{0.85\columnwidth}{!}{%
		\begin{tabular}{lcccc}
\toprule
&Conv-1&Conv-5&HOG&CN\\\midrule
Feature dimension, $D$&96&512&31&11\\
Filter dimension, $C$&16&64&10&3\\\bottomrule
\end{tabular}
	}\vspace{1mm}%
	\caption{The settings of the proposed factorized convolution approach, as employed in our experiments. For each feature, we show the dimensionality $D$ and the number of filters $C$.}%
	\label{tab:fac_settings}%
	\vspace{-3mm}
\end{table}

\subsection{Implementation Details}
\label{sec:details}
Our tracker is implemented in Matlab. We apply the same feature representation as C-COT, namely a combination of the first (Conv-1) and last (Conv-5) convolutional layer in the VGG-m network \cite{Chatfield14}, along with HOG \cite{Dalal05} and Color Names (CN) \cite{Weijer09a}. For the factorized convolution presented in section~\ref{sec:fac}, we learn one coefficient matrix $P$ for each feature type. The settings for each feature is summarized in table~\ref{tab:fac_settings}. The regularization parameter $\lambda$ in \eqref{eq:loss_fac} is set to $2 \cdot 10^{-7}$. The loss \eqref{eq:loss_fac} is optimized in the first frame using 10 Gauss-Newton iterations and 20 CG iterations for the subproblems \eqref{eq:GN_loss}. In the first iteration $i=0$, the filter $\hat{f}_0$ is initialized to zero. To preserve the deterministic property of the tracker, we initialize the coefficient matrix $P_0$ by PCA, though we found random initialization to be equally robust.

For the sample space model, presented in section~\ref{sec:sample_model}, we set the learning rate to $\gamma = 0.012$. The number of components are set to $L = 50$, which represents an 8-fold reduction compared to the number of samples ($M=400$) used in C-COT. We update the filter in every $N_\text{S} = 6$ frame (section~\ref{sec:optimization}). We use the same number of $N_\text{CG} = 5$ Conjugate Gradient iterations as in C-COT. Note that \emph{all} parameters settings are kept fixed for all videos in a dataset.

\begin{table}[!t]
	\centering
	\resizebox{\columnwidth}{!}{%
		\begin{tabular}{l@{}c@{}c@{}c@{}c@{}c@{}c@{\hspace{-3mm}}c}
\toprule
&Baseline&&Factorized&&Sample&&Model\\
&C-COT&$\implies$&Convolution&$\implies$&Space Model&$\implies$&Update\\
&(Sec.~\ref{sec:CCOT})&&(Sec.~\ref{sec:fac})&&(Sec.~\ref{sec:sample_model})&&(Sec.~\ref{sec:optimization})\\\midrule
EAO&0.331&&0.342&&0.352&&\tabfirst{0.374}\\
FPS&0.3&&1.1&&2.6&&6.0\\
Compl.\ change&-&&$D \rightarrow C$&&$M \rightarrow L$&&$N_\text{CG} \rightarrow \frac{N_\text{CG}}{N_\text{S}}$\\
Compl.\ red.&-&&$6 \times$&&$8 \times$&&$6 \times$\\
\bottomrule
\end{tabular}

	}\vspace{1mm}%
	\caption{Analysis of our approach on the VOT2016. The impact of progressively integrating one contribution at the time, from left to right, is displayed. We show the performance in Expected Average Overlap (EAO) and speed in FPS (benchmarked on a single CPU). We also summarize the reduction in learning complexity $\ordo(N_\text{CG} D M \bar{K})$ obtained in each step, both symbolically and in absolute numbers (bottom row) using our settings. Our contributions systematically improve both performance and speed.}%
	\label{tab:baseline}%
	\vspace{-4mm}
\end{table}

\subsection{Baseline Comparison}
\label{sec:baseline_results}

Here, we analyze our approach on the VOT2016 benchmark by demonstrating the impact of progressively integrating our contributions. The VOT2016 dataset consists of 60 videos compiled from a set of more than 300 videos. The performance is evaluated both in terms of accuracy (average overlap during successful tracking) and robustness (failure rate). The overall performance is evaluated using Expected Average Overlap (EAO) which accounts for both accuracy and robustness. We refer to \cite{VOT2015} for details.

Table~\ref{tab:baseline} shows an analysis of our contributions. The integration of our factorized convolution into the baseline leads to a performance improvement and a significant reduction in complexity ($6 \times$). The sample space model further improves the performance by a relative gain of $2.9 \%$ in EAO, while reducing the learning complexity by a factor of $8$. Additionally incorporating our proposed model update elevates us to an EAO score of $0.374$, leading to a final relative gain of $13.0 \%$ compared to the baseline. In table~\ref{tab:baseline} we also show the impact on the tracker speed achieved by our contributions. For a fair comparison, we report the FPS measured on a single CPU for all entries in the table, without accounting for feature extraction time. Each of our contributions systematically improves the speed of the tracker, combining to a 20-fold final gain compared to the baseline. When including all steps (also feature extraction), the GPU version of our tracker operates at 8 FPS.

\begin{figure}[!t]
	\centering%
	\newcommand{\wid}{0.8\columnwidth}
	\includegraphics[width = \wid]{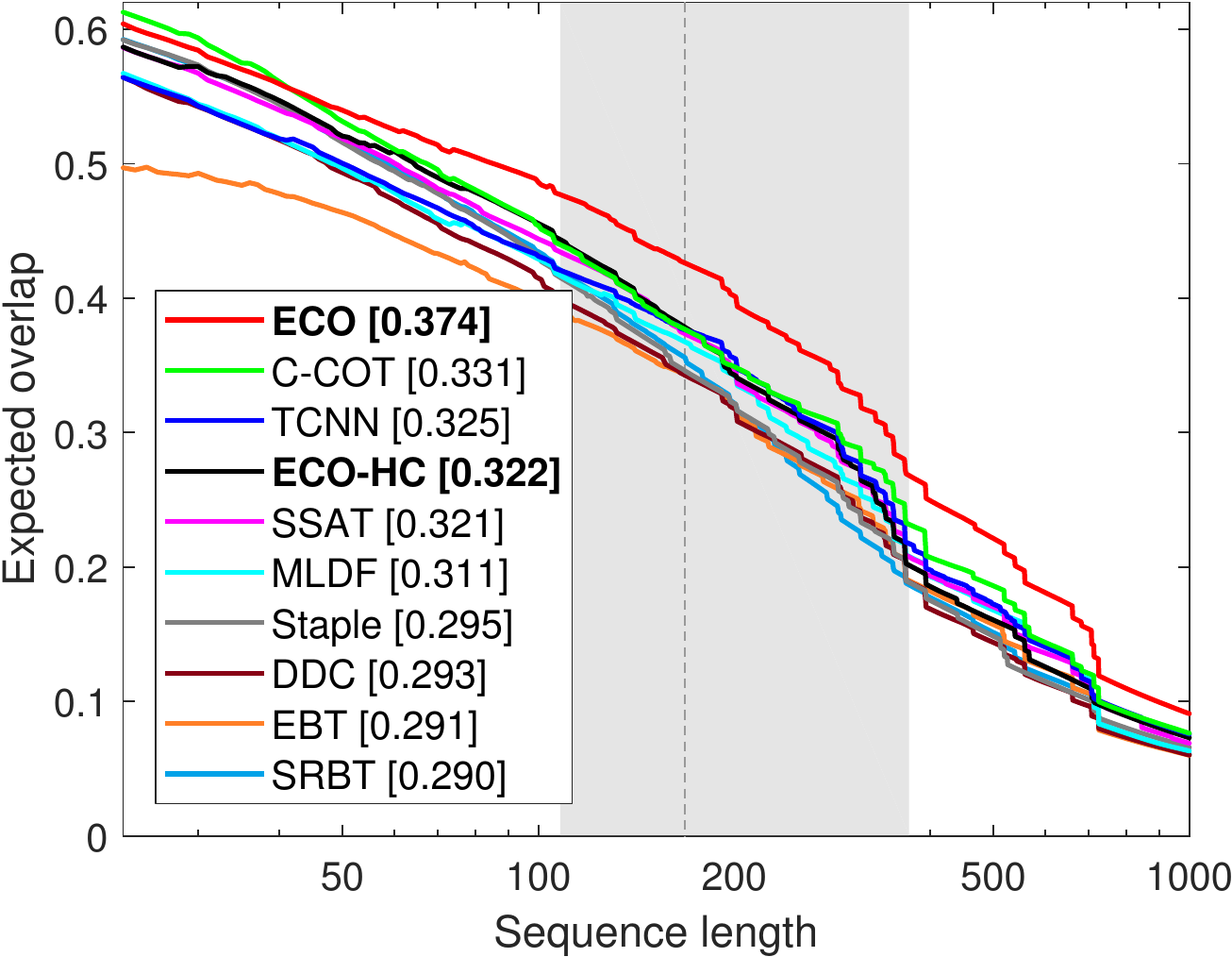}\vspace{-0.5mm}
	\caption{Expected Average Overlap (EAO) curve on VOT2016. Only the top 10 trackers are shown for clarity. The EAO measure, computed as the average EAO over typical sequence lengths (grey region), is displayed in the legend (see \cite{VOT2015} for details).
	%Our approach achieves outstanding performance with a relative gain of $13.0\%$ in the EAO measure compared to the top performer in the challenge.
	}\vspace{-2mm}%
	\label{fig:vot_eao}
\end{figure}

We found the settings in table~\ref{tab:fac_settings} to be insensitive to minor changes. Substantial gain in speed can be obtained by reducing the number of filters $C$, at the cost of a slight reduction in performance. To further analyze the impact of our jointly learned factorized convolution approach, we compare with applying PCA in the first frame to obtain the matrix $P$. PCA degrades the EAO from $0.331$ to $0.319$, while our discriminative learning based method achieves $0.342$.

We observed that our sample model provides consistently better results compared to the training sample set management employed in C-COT when using the same number of components and samples ($L = M$). This is particularly evident for a smaller number of components/samples: When reducing the number of samples from $M = 400$ to $M = 50$ in the standard approach, the EAO decreases from $0.342$ to $0.338$ ($- 1.2 \%$). Instead, when using our approach with $L = 50$ components, the EAO increases by $+ 2.9 \%$ to $0.351$. In case of the model update, we observed an upward trend in performance when increasing $N_\text{S}$ from 1 to 6. When increasing $N_\text{S}$ further, a gradual downward trend was observed. We therefore use $N_\text{S} = 6$ throughout our experiments.

\begin{table}[!t]
	\centering
	\resizebox{1.01\columnwidth}{!}{%
		\begin{tabular}{l@{~}c@{~~}c@{~~}c@{~~}c@{~~}c@{~~}c@{~~}c@{~~}c@{~~}c@{~~}c}
	\toprule
	&SRBT&EBT&DDC&Staple&MLDF&SSAT&TCNN&C-COT&\textbf{ECO-HC}&\textbf{ECO}\\
	&\cite{VOT2016}&\cite{EBT}&\cite{VOT2016}&\cite{Staple}&\cite{VOT2016}&\cite{VOT2016}&\cite{TCNN}&\cite{DanelljanECCV2016}&Ours&Ours\\\midrule
	EAO&0.290&0.291&0.293&0.295&0.311&0.321&0.325&\tabsecond{0.331}&0.322&\tabfirst{0.374}\\
	Fail.\ rt.&1.25&0.90&1.23&1.35&\tabsecond{0.83}&1.04&0.96&0.85&1.08&\tabfirst{0.72}\\
	Acc.&0.50&0.44&0.53&\tabsecond{0.54}&0.48&\tabfirst{0.57}&\tabsecond{0.54}&0.52&0.53&\tabsecond{0.54}\\
	EFO&3.69&3.01&0.20&\tabsecond{11.14}&1.48&0.48&1.05&0.51&\tabfirst{15.13}&4.53\\\bottomrule
\end{tabular}

%\begin{tabular}{l@{~}c@{~~}c@{~~}c@{~~}c@{~~}c@{~~}c@{~~}c@{~~}c@{~~}c@{~~}c@{~~}c@{~~}c}
%\toprule
%&DNT&Staple+&SRBT&EBT&DDC&Staple&MLDF&SSAT&TCNN&C-COT&\textbf{ECO-HC}&\textbf{ECO}\\
%&\cite{DNT}&\cite{VOT2016}&\cite{VOT2016}&\cite{EBT}&\cite{VOT2016}&\cite{Staple}&\cite{VOT2016}&\cite{VOT2016}&\cite{TCNN}&\cite{DanelljanECCV2016}&Ours&Ours\\\midrule
%EAO&0.278&0.286&0.290&0.291&0.293&0.295&0.311&0.321&0.325&\tabsecond{0.331}&-&\tabfirst{0.374}\\
%Fail.\ rt.&1.18&1.32&1.25&0.90&1.23&1.35&\tabsecond{0.83}&1.04&0.96&0.85&-&\tabfirst{0.72}\\
%Acc.&0.50&\tabsecond{0.55}&0.50&0.44&0.53&0.54&0.48&\tabfirst{0.57}&0.54&0.52&-&0.54\\
%EFO&1.13&\tabfirst{44.77}&3.69&3.01&0.20&\tabsecond{11.14}&1.48&0.48&1.05&0.51&-&4.53\\\bottomrule
%\end{tabular}

	}\vspace{1mm}%
	\caption{State-of-the-art in terms of expected average overlap (EAO), robustness (failure rate), accuracy, and speed (in EFO units) on the VOT2016 dataset. Only the top-10 trackers are shown. Our deep feature based ECO achieve superior EAO, while our hand-crafted feature version (ECO-HC) has the best speed.}%
	\label{tab:VOT_sota}%
	\vspace{-2mm}
\end{table}

\subsection{State-of-the-art Comparison}
Here, we compare our approach with state-of-the-art trackers on four challenging tracking benchmarks. Detailed results are provided in the supplementary material.

\begin{figure*}[!t]
	\centering%
	\newcommand{\wid}{0.33\textwidth}\vspace{-3.5mm}
	\subfloat[UAV123\label{fig:sota_uav}]{\includegraphics[width = \wid]{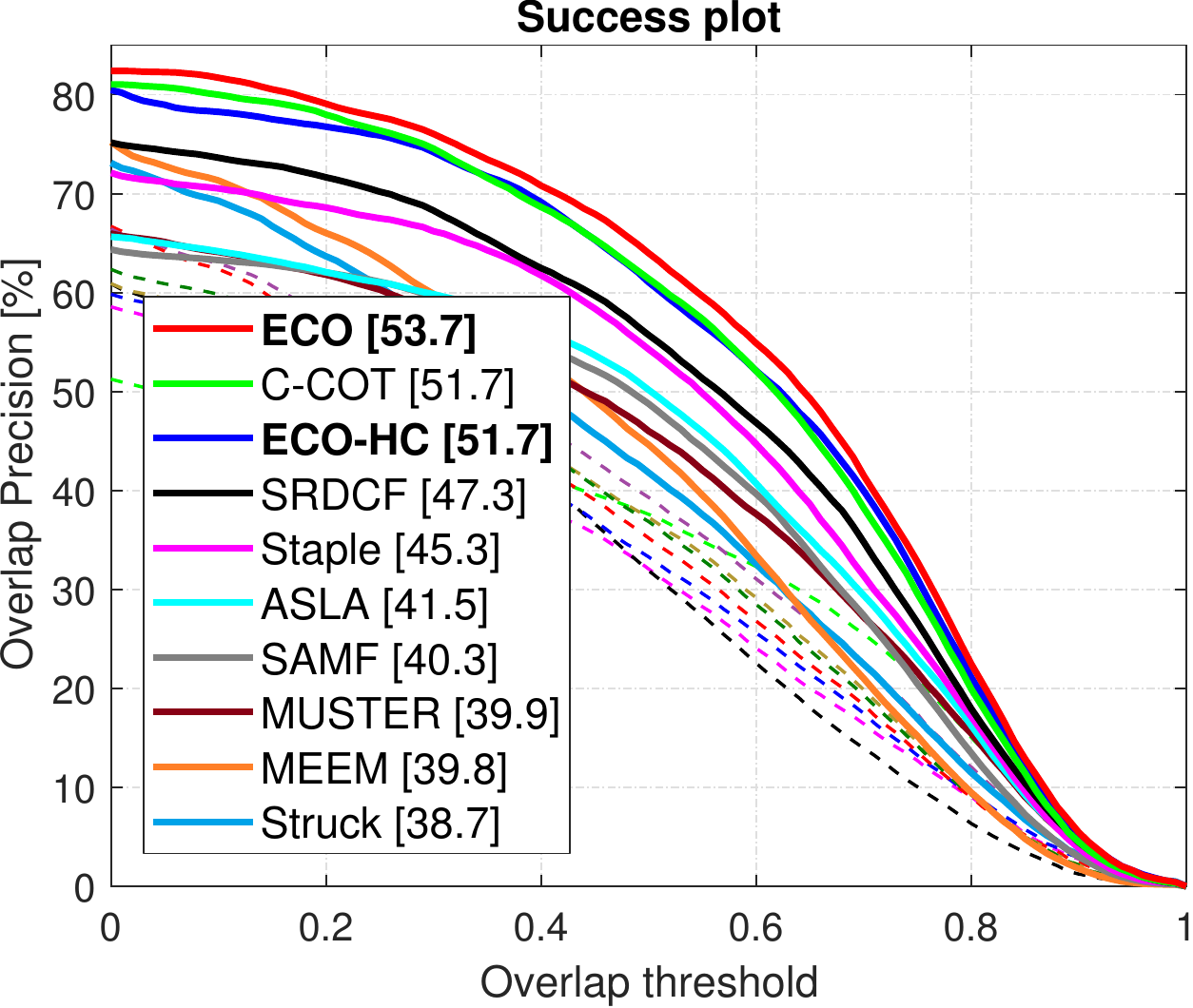}}%
	\subfloat[OTB-2015\label{fig:sota_otb}]{\includegraphics[width = \wid]{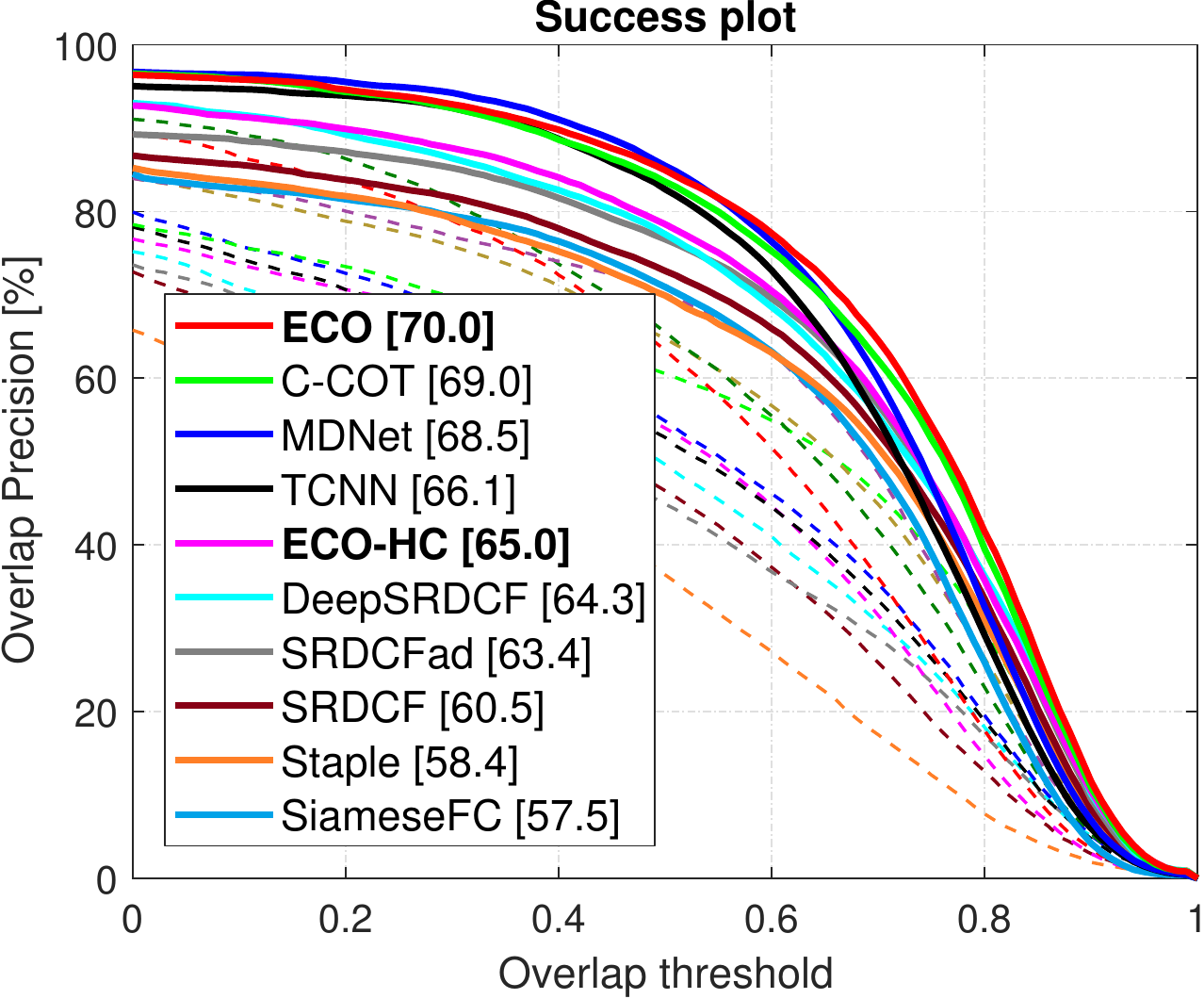}}%
	\subfloat[Temple-Color\label{fig:sota_tpl}]{\includegraphics[width = \wid]{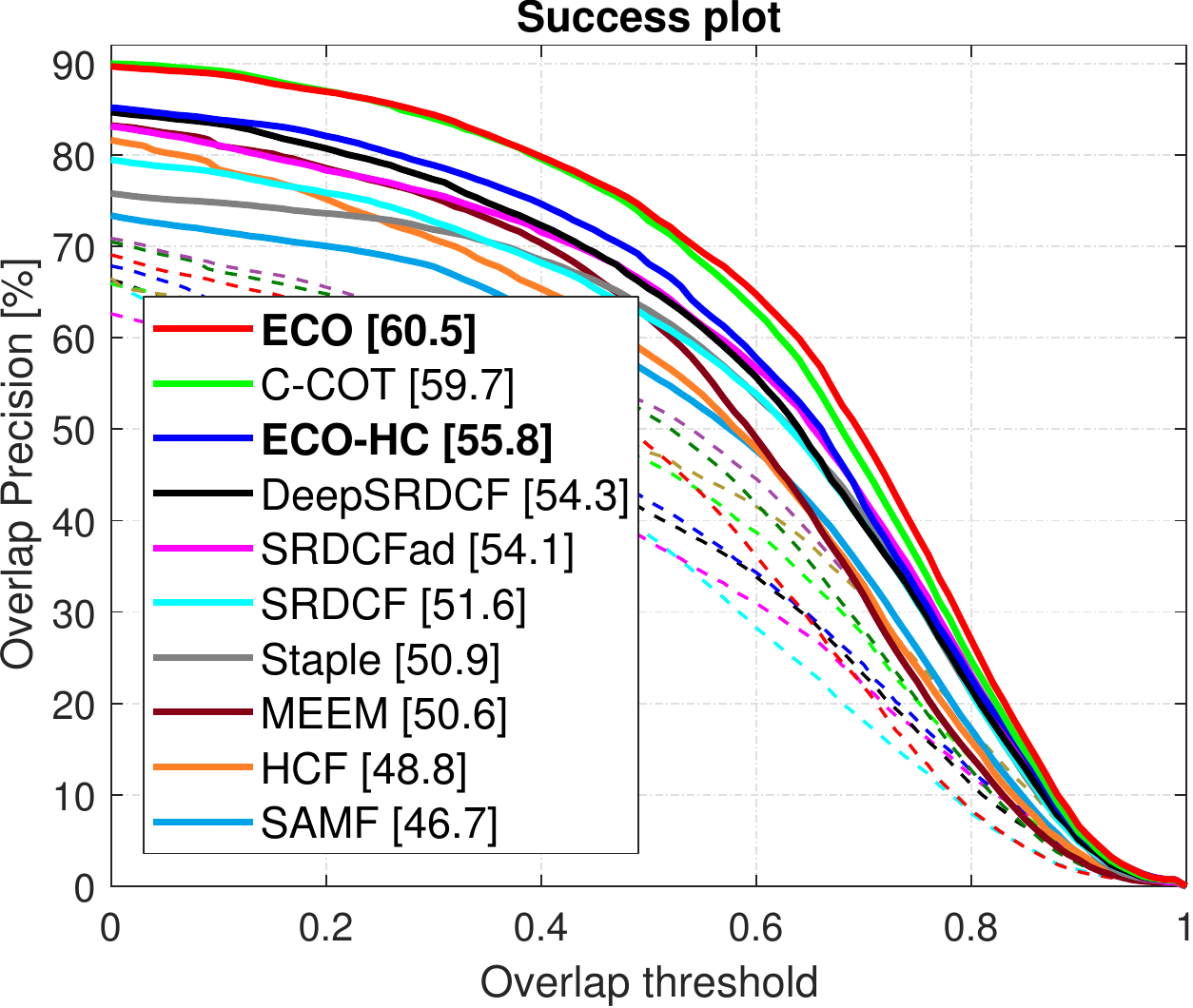}}%
	\caption{Success plots on the UAV-123 (a), OTB-2015 (b) and TempleColor (c) datasets. Only the top 10 trackers are shown in the legend for clarity. The AUC score of each tracker is shown in the legend. Our approach significantly improves the state-of-the-art on all datasets.}\vspace{-2mm}%
	\label{fig:success_plots}
\end{figure*}

\parsection{VOT2016 Dataset}
In table~\ref{tab:VOT_sota} we compare our approach, in terms of expected average overlap (EAO), robustness, accuracy and speed (in EFO units), with the top-ranked trackers in the VOT2016 challenge. The first-ranked performer in VOT2016 challenge, C-COT, provides an EAO score of $0.331$. Our approach achieves a relative gain of $13.0\%$ in EAO compared to C-COT. Further, our ECO tracker achieves the best failure rate of $0.72$ while maintaining a competitive accuracy. We also report the total speed in terms of EFO, which normalizes the speed with respect to hardware performance. Note that EFO also takes feature extraction time into account, a major additive complexity that is independent of our DCF improvements. In the comparison, our tracker ECO-HC using only hand-crafted features (HOG and Color Names) achieves the best speed. Among the top three trackers in the challenge, which are all based on deep features, TCNN \cite{TCNN} obtains the best speed with an EFO of $1.05$. Our deep feature version (ECO) achieves an almost 5-fold speedup in EFO and a relative performance improvement of $15.1 \%$ in EAO compared to TCNN. Figure~\ref{fig:vot_eao} displays the EAO curves of the top-10 trackers.

\parsection{UAV123 Dataset}
Aerial tracking using unmanned aerial vehicles (UAVs) has received much attention recently, with many vision applications, including wild-life monitoring, search and rescue, navigation, and crowd surveillance. In these applications, persistent UAV navigation is required, for which real-time tracking output is crucial. In such cases, the desired tracker should be accurate and robust, while operating in real-time under limited hardware capabilities, \eg, CPUs or mobile GPU platforms. We therefore introduce a real-time variant of our method (ECO-HC), based on hand-crafted features (HOG and Color Names), operating at 60 FPS on a single i7 CPU (including feature extraction).

We evaluate our trackers on the recently introduced aerial video benchmark, UAV123 \cite{UAV123}, for low altitude UAV target tracking. The dataset consists of 123 aerial videos with more than 110K frames. The trackers are evaluated using success plot \cite{Wu13}, calculated as percentage of frames with an intersection-over-union (IOU) overlap exceeding a threshold. Trackers are ranked using the area-under-the-curve (AUC) score. Figure~\ref{fig:sota_uav} shows the success plot over all the 123 videos in the dataset. We compare with all tracking results reported in \cite{UAV123} and further add Staple \cite{Staple}, due to its high frame-rate, and C-COT \cite{DanelljanECCV2016}. Among the top 5 compared trackers, only Staple runs at real-time, with an AUC score of $45.3\%$. Our ECO-HC tracker also operates in real-time (60 FPS), with an AUC score of $51.7\%$, significantly outperforming Staple by $6.4\%$. C-COT obtains an AUC score of $51.7\%$. Our ECO outperforms C-COT, achieving an AUC score of $53.7\%$, using same features.

\parsection{OTB2015 Dataset}
We compare our tracker with 20 state-of-the-art methods: TLD \cite{Mikolajczyk10d}, Struck \cite{Torr11b}, CFLB \cite{GaloogahiCVPR2015}, ACT \cite{DanelljanCVPR14}, TGPR \cite{TGPR2014}, KCF \cite{Henriques14}, DSST \cite{DanelljanBMVC14}, SAMF \cite{Li2014}, MEEM \cite{MEEM2014}, DAT \cite{possegger15a}, LCT \cite{LTC_CVPR15}, HCF \cite{HCF_ICCV15}, SRDCF \cite{DanelljanICCV2015}, SRDCFad \cite{DanelljanCVPR2016a}, DeepSRDCF \cite{DanelljanVOT2015}, Staple \cite{Staple}, MDNet \cite{MDNet}, SiameseFC \cite{SiameseFC}, TCNN \cite{TCNN} and C-COT \cite{DanelljanECCV2016}.

Figure~\ref{fig:sota_otb} shows the success plot over all the 100 videos in the OTB-2015
dataset \cite{OTB2015}. Among the compared trackers using hand-crafted features, SRDCFad provides the best results with an AUC score of $63.4\%$. Our proposed method, ECO-HC, also employing hand-crafted features outperforms SRDCFad with an AUC score of $65.0\%$, while running on a CPU with a speed of 60 FPS. Among the compared deep feature trackers, C-COT, MDNet and TCNN provide the best results with AUC scores of $69.0\%$, $68.5\%$ and $66.1\%$ respectively. Our approach ECO, provides the best performance with an AUC score of $70.0\%$.

\parsection{TempleColor Dataset}
In figure~\ref{fig:sota_tpl} we present results on the TempleColor dataset \cite{TempleColor} containing 128 videos. Our method again achieves a substantial improvement over C-COT, with a gain of $0.8 \%$ in AUC.

\section{Conclusions} 
We revisit the core DCF formulation to counter the issues of over-fitting and computational complexity. 
We introduce a factorized convolution operator to reduce the number of parameters in the model. We also propose a compact generative model of the training sample distribution to drastically reduce memory and time complexity of the learning, while enhancing sample diversity. Lastly, we suggest a simple yet effective model update strategy that reduces over-fitting to recent samples. Experiments on four datasets demonstrate state-of-the-art performance with improved frame rate. 

\noindent\textbf{Acknowledgments}:
This work has been supported by SSF (SymbiCloud), VR (EMC${}^2$, starting grant 2016-05543), SNIC, WASP, Visual Sweden, and Nvidia.

{\small
\bibliographystyle{ieee}
\bibliography{references}
}

\clearpage

\section*{Supplementary Material}

This supplementary material contains additional details and derivations related to the our approach presented in section~\ref{sec:method}. It also includes hardware specifications and additional experimental results on the VOT2016 and OTB-2015 datasets.

\subsection*{Complexity Analysis of the Learning}
\label{sec:supp_complexity}

Here, we derive the computational complexity of the learning step in the baseline C-COT \cite{DanelljanECCV2016}. The learning itself is completely dominated by the problem of solving the normal equations \eqref{eq:normal_eq}.
This linear system is iteratively solved using the Conjugate Gradient (CG) method \cite{NumericalOptimization,CGpain}. The dominating computation in CG is the evaluation of the left-hand side of \eqref{eq:normal_eq}, which is performed once per CG iteration. This computation is performed as
\begin{equation}
\label{eq:cg_order}
A\ctp (\Gamma (A \vecft{f})) + W\ctp (W \vecft{f}),
\end{equation}
where the parentheses are used to indicate the order in which the operations are performed. Since the conjugate symmetry in the filter $\vecft{f}$ is preserved by the operations in \eqref{eq:cg_order}, only half of the spectrum needs to be processed. We can therefore regard $\vecft{f}$ as a complex vector of $\sum_d K_d = D \bar{K}$ elements, where $K_d$ is the bandwidth of channel $d$ in the filter (see section~\ref{sec:CCOT}), $\bar{K} = \frac{1}{D} \sum_d K_d$ is the average number of Fourier coefficients per channel and $D$ is the number of feature channels $d$.

The matrix $A$ contains a diagonal block $A_{j,d}$ of size $K \times K_d$ for each sample $j \in \{1, \ldots, M\}$ and channel $d \in \{1, \ldots, D\}$. Here, we have defined $K = \max_d K_d$. The diagonal of $A_{j,d}$ consists of the elements $\{X^d_j[k] \hat{b}_d[k]\}_{k=0}^{K_d}$. As previously shown for the discrete DCF case \cite{galoogahiICCV13}, the matrix $A$ can be permuted to a block diagonal matrix $\tilde{A} = \oplus_{k=0}^K \tilde{A}_k$, where $\tilde{A}_k$ contains the elements $(\tilde{A}_k)_{j,d} = X^d_j[k] \hat{b}_d[k]$. The operations $\vecft{f} \mapsto A \vecft{f}$ and $\vecft{v} \mapsto A\ctp \vecft{v}$ can thus be implemented as block-wise dense matrix-vector multiplications, with a total of $\ordo(D M \bar{K})$ operations. Moreover, $\Gamma$ is a diagonal matrix containing the weights $\alpha_j$, giving $\ordo(M \bar{K})$ operations.

In the second term of \eqref{eq:cg_order}, arising from the spatial regularization in the loss \eqref{eq:loss_spatial}, $W$ and $W\ctp$ are convolution matrices with the kernel $\hat{w}[k]$. These operations have a complexity of $\ordo(D \bar{K} K_w)$, where $K_w$ are the number of non-zero coefficients in $\hat{w}$ (\ie the size of the kernel). In practice however, the kernel $\hat{w}[k]$ is small (typically $5 \times 5$), having a lesser impact on the overall complexity. To simplify the complexity expression, we therefore disregard this term. By taking the number of CG iterations $N_\text{CG}$ into account, we thus obtain the final expression  $\ordo(N_\text{CG} D M \bar{K})$ for the complexity of the learning step.

The preprocessing steps needed for the CG optimization only have a marginal impact on the overall learning time. The most significant of these being the Fast Fourier Transform (FFT) of the feature map, having a $\ordo( \sum_d N_d \log N_d)$ complexity, where $N_d$ is the resolution of feature channel $d$. But since the FFT computations correspond to roughly $1 \%$ of the total time in C-COT, we exclude this part.

\subsection*{Factorized Convolution Operator Optimization}
\label{sec:supp_optimization}

Here, we provide more details regarding the optimization procedure for learning the factorized convolution operator (section~\ref{sec:fac}). We consider the case of learning the factorized operator $S_{Pf}\{x\}$ in \eqref{eq:fac_conv_op} based on a single sample $(x,y)$,
\begin{equation}
\label{eq:fac_spatial}
E(f,P) = \left\| S_{f,P}\{x\} - y \right\|^2_{L^2} + \sum_{c=1}^{C} \left\| w f^c \right\|^2_{L^2} + \lambda \|P\|_F^2 \,.
\end{equation}
The loss is obtained by employing the factorized operator $S_{f,P}\{x\}$ in the data term of the original loss \eqref{eq:loss_spatial} and adding a regularization on the Frobenius norm $\|P\|_F^2$ of $P$.

By applying the Parseval's formula to the first two terms of \eqref{eq:fac_spatial} and utilizing the linearity and convolution properties of the Fourier series coefficients, we obtain the equivalent loss \eqref{eq:loss_fac}, where we have defined the interpolated feature map as $z = J\{x\}$ to simplify notation. Note that the matrix-vector products in \eqref{eq:loss_fac} are performed point-wise,
\begin{equation}
	\label{eq:prod}
	(\hat{z}\tp P \hat{f})[k] = \sum_{d=1}^{D} \sum_{c=1}^{C} \hat{z}^d[k] p_{d,c} \hat{f^c}[k] \,, \quad k \in \integers  \,.
\end{equation}

We use the Gauss-Newton method \cite{NumericalOptimization} to optimize the non-linear least squares problem \eqref{eq:loss_fac}. In each iteration $i$, the residual in the data-term is linearized by performing a first order Taylor expansion \eqref{eq:taylor} at the current estimate $(\hat{f}_i, P_i)$. This gives the following quadratic sub-problem \eqref{eq:GN_loss}.
To derive a simple formula for the normal equations of \eqref{eq:GN_loss}, we first introduce some notation. Let $\vecft{f}$ be the vectorization of $\hat{f}_{i,\Delta}$, analogously to \eqref{eq:normal_eq}, and define $\mathbf{\Delta p} = \vecop(\Delta P)$. Further, let $\mathbf{p}_c$ denote the $c$th column in $P_i$ and set $\mathbf{p} = \vecop(P_i)$. We then define the matrices,
\begin{align}
	\label{eq:notation}
	A_P \!=& \! \begin{bmatrix}
		\zmat{K-K_1}{2K_1+1} & \!\cdots\! & \zmat{K-K_C}{2K_C+1} \vspace{2mm}\\
		\diag \begin{pmatrix} \hat{z}[-K_1]\tp \mathbf{p}_1 \\ \vdots \\ \hat{z}[K_1]\tp \mathbf{p}_1 \end{pmatrix} &
		\!\cdots\! &
		\diag \begin{pmatrix} \hat{z}[-K_C]\tp \mathbf{p}_C \\ \vdots \\ \hat{z}[K_C]\tp \mathbf{p}_C \end{pmatrix} \vspace{2mm}\\
		\zmat{K-K_1}{2K_1+1} & \!\cdots\! & \zmat{K-K_C}{2K_C+1}
	\end{bmatrix}
	\nonumber \\
	B_f =& \begin{pmatrix}
		(\hat{f}_i \otimes \hat{z})[-K]\tp \\ \vdots \\ (\hat{f}_i \otimes \hat{z})[K]\tp
	\end{pmatrix}
	\, .
\end{align}
Here, $A_P$ has a structure very similar to the matrix $A$ in \eqref{eq:normal_eq}, but contains only a single training sample. Note that the diagonal blocks in $A_P$ are padded with zero matrices $\zmat{M}{N}$ along the columns to achieve the same number of $2K+1$ rows.

The Gauss-Newton sub-problem \eqref{eq:GN_loss} can then be expressed as,
\begin{equation}
	\label{eq:GN_quadratic}
	\tilde{E}(\vecft{f}, \!\mathbf{\Delta p}) \!=\! \left\| A_P \vecft{f} \!+\! B_f \mathbf{\Delta p} \!-\! \vecft{y} \right\|_2^2\! + \big\| W \vecft{f} \big\|^2_2 + \lambda \!\left\| \mathbf{p} \!+\! \mathbf{\Delta p} \right\|_2^2.
\end{equation}
Here, the convolution matrix $W$ and the vectorization $\vecft{y}$ are defined as in \eqref{eq:normal_eq}. The normal equations of \eqref{eq:GN_quadratic} are obtained by setting the gradient to zero,
\begin{equation}
	\label{eq:GN_normal}
	\begin{bmatrix}
		A_P\ctp A_P + W\ctp W & A_P\ctp B_f \\ B_f\ctp A_P & B_f\ctp B_f + \lambda I
	\end{bmatrix}\!
	\begin{bmatrix}
		\vecft{f} \\ \mathbf{\Delta p}
	\end{bmatrix}
	=
	\begin{bmatrix}
		A_P\ctp \vecft{y} \\ B_f\ctp \vecft{y} - \lambda \mathbf{p}
	\end{bmatrix}
	\!.
\end{equation}
We employ the Conjugate Gradient method to iteratively solve the sub-problem \eqref{eq:GN_normal}.

\subsection*{Hardware Specifications}
\label{sec:supp_hw}

Our tracker is implemented in Matlab and uses Matconvnet \cite{matconvnet} for deep feature extraction. The frame-rate measurements of our CPU implementation were performed on a desktop computer with a 4-core Intel Core i7-6700 CPU at $3.4$ GHz. The frame-rate measurements of our GPU implementation were performed on a Tesla K40 GPU.

\subsection*{Additional Results on VOT2016}
\label{sec:supp_VOT}
Here, we provide further experimental evaluation on the VOT2016 dataset \cite{VOT2016} with 60 videos. The videos and the evaluation toolkit can be obtained from \url{http://www.votchallenge.net/vot2016/}.

\begin{figure}[!t]
	\centering
	\newcommand{\wid}{\columnwidth}
	\newcommand{\name}{figures/sota_VOT}
	\includegraphics[width=\wid]{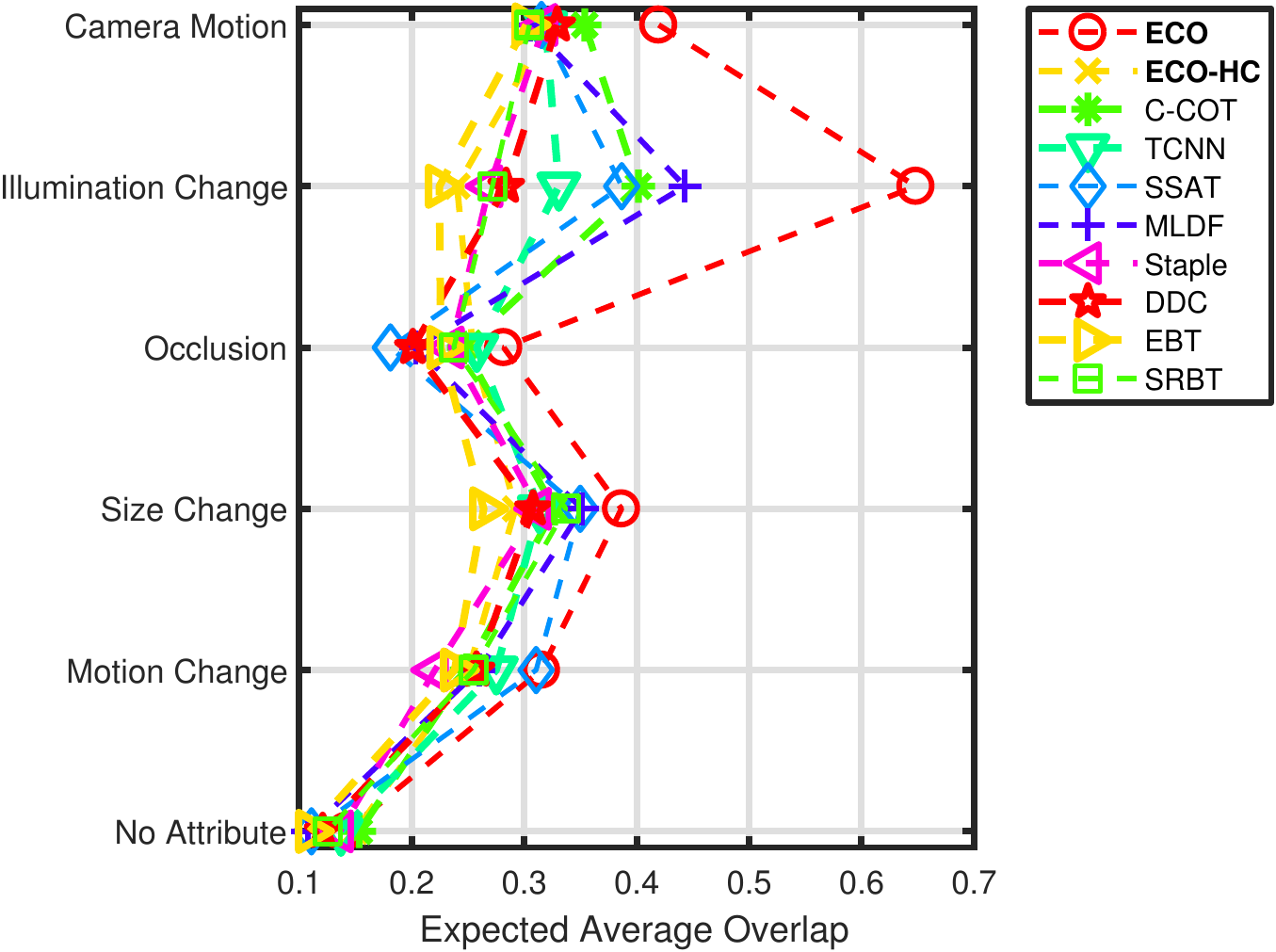}%
	\caption{Expected Average Overlap (EAO) scores for each attribute on the VOT2016 dataset. Here, \emph{empty} denotes frames with no labeled attribute.}
	\label{fig:VOT_attribute}
\end{figure}

In the VOT2016 dataset, each frame is labeled with five different attributes: camera motion, illumination change, occlusion, size change and motion change. Figure~\ref{fig:VOT_attribute} visualizes the EAO of each attribute individually. Our approach achieves the best results on three attributes and improves over the baseline C-COT \cite{DanelljanECCV2016} on all five attributes.

\subsection*{Additional Results on OTB-2015}
\label{sec:supp_OTB}
Here, we report additional results on the OTB-2015 dataset \cite{OTB2015} with 100 videos. The ground truth annotations and videos are available at \url{https://sites.google.com/site/benchmarkpami/}.

\begin{figure*}[!t]
	\centering
	\newcommand{\wid}{0.33\textwidth}
	\newcommand{\name}{figures/sota_OTB100}
	\newcommand{\eval}{OPE}
	\includegraphics[width=\wid]{\name /quality_plot_overlap_\eval _AUC.pdf}%
	\includegraphics[width=\wid]{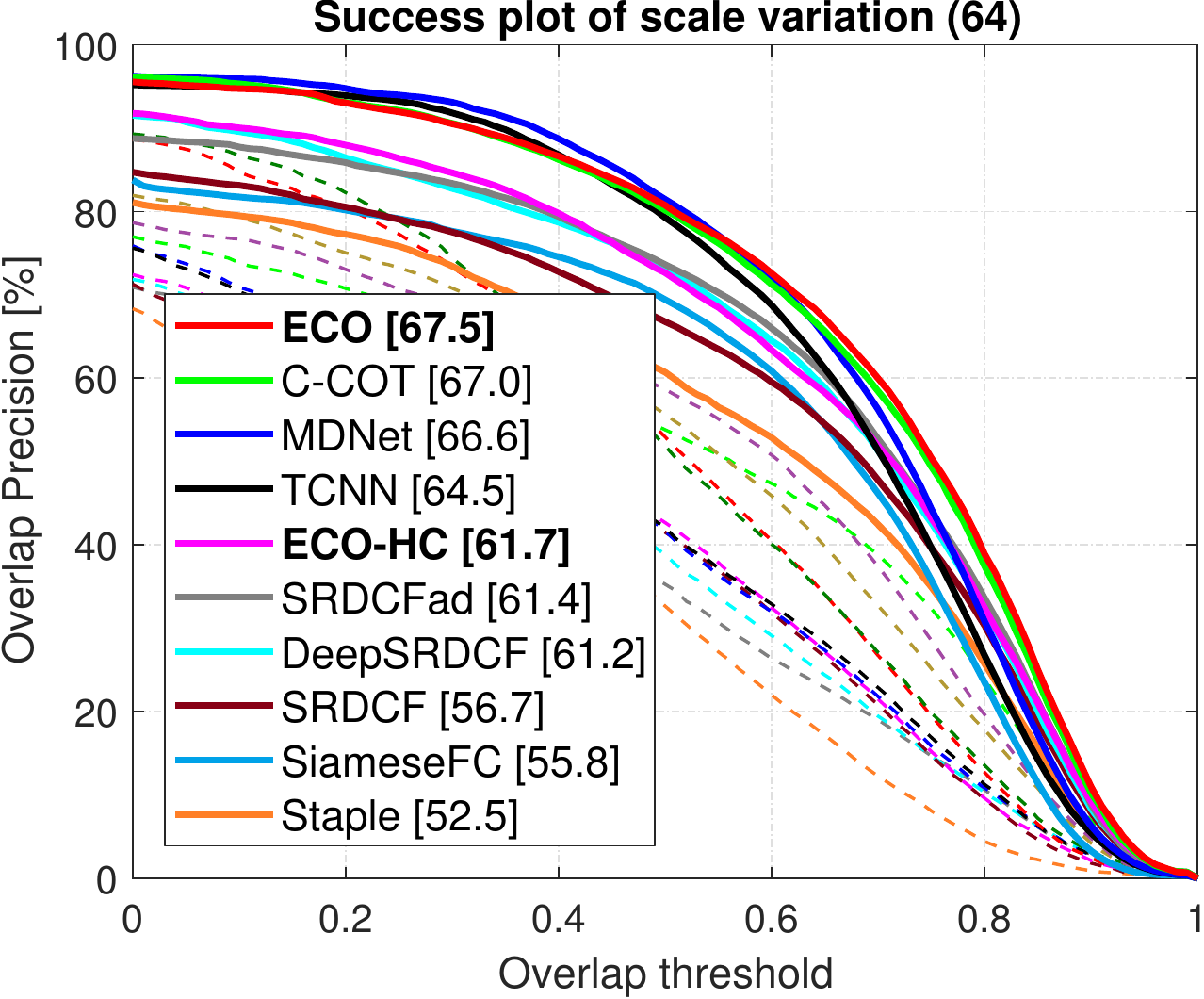}%
	\includegraphics[width=\wid]{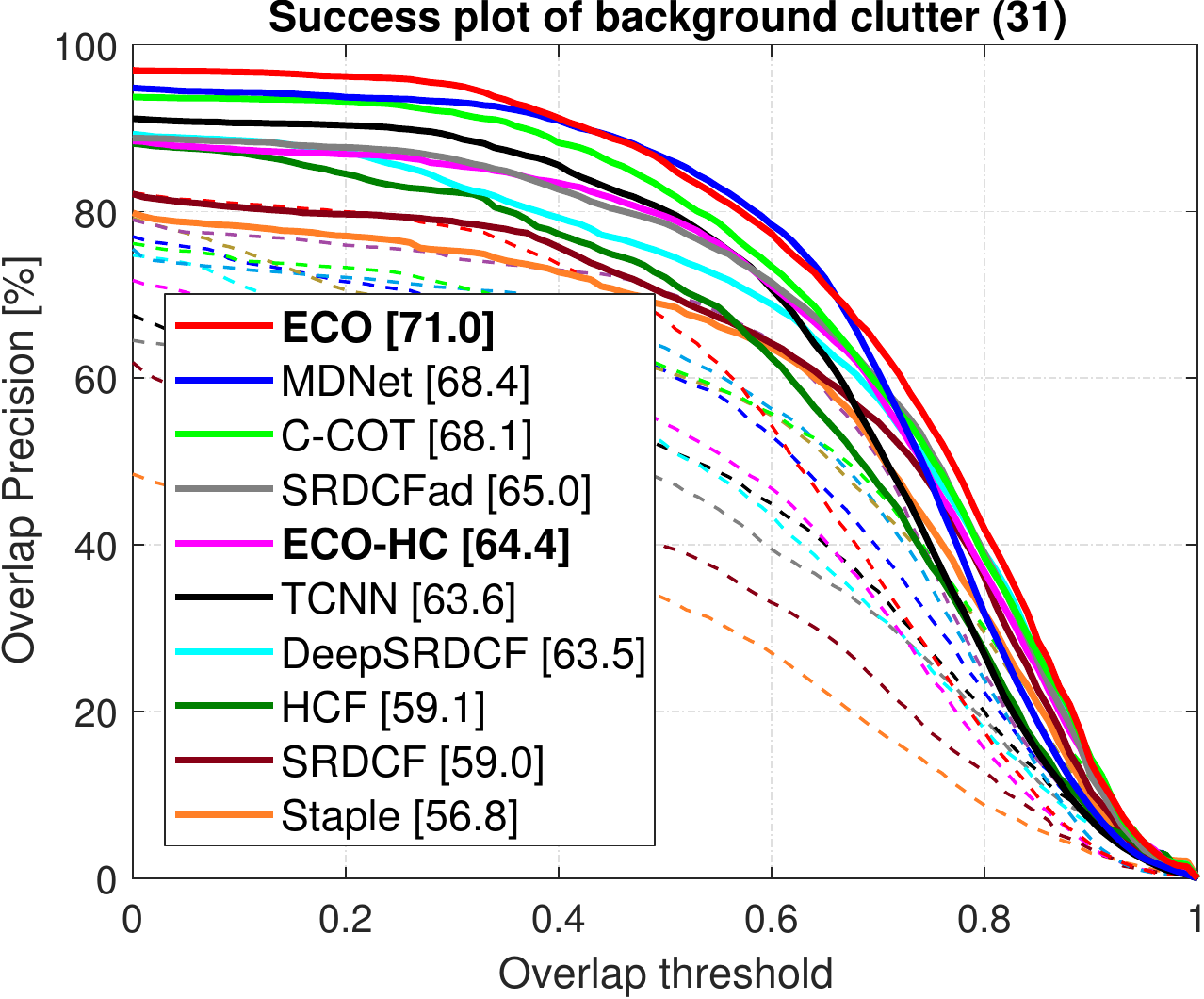}
	\includegraphics[width=\wid]{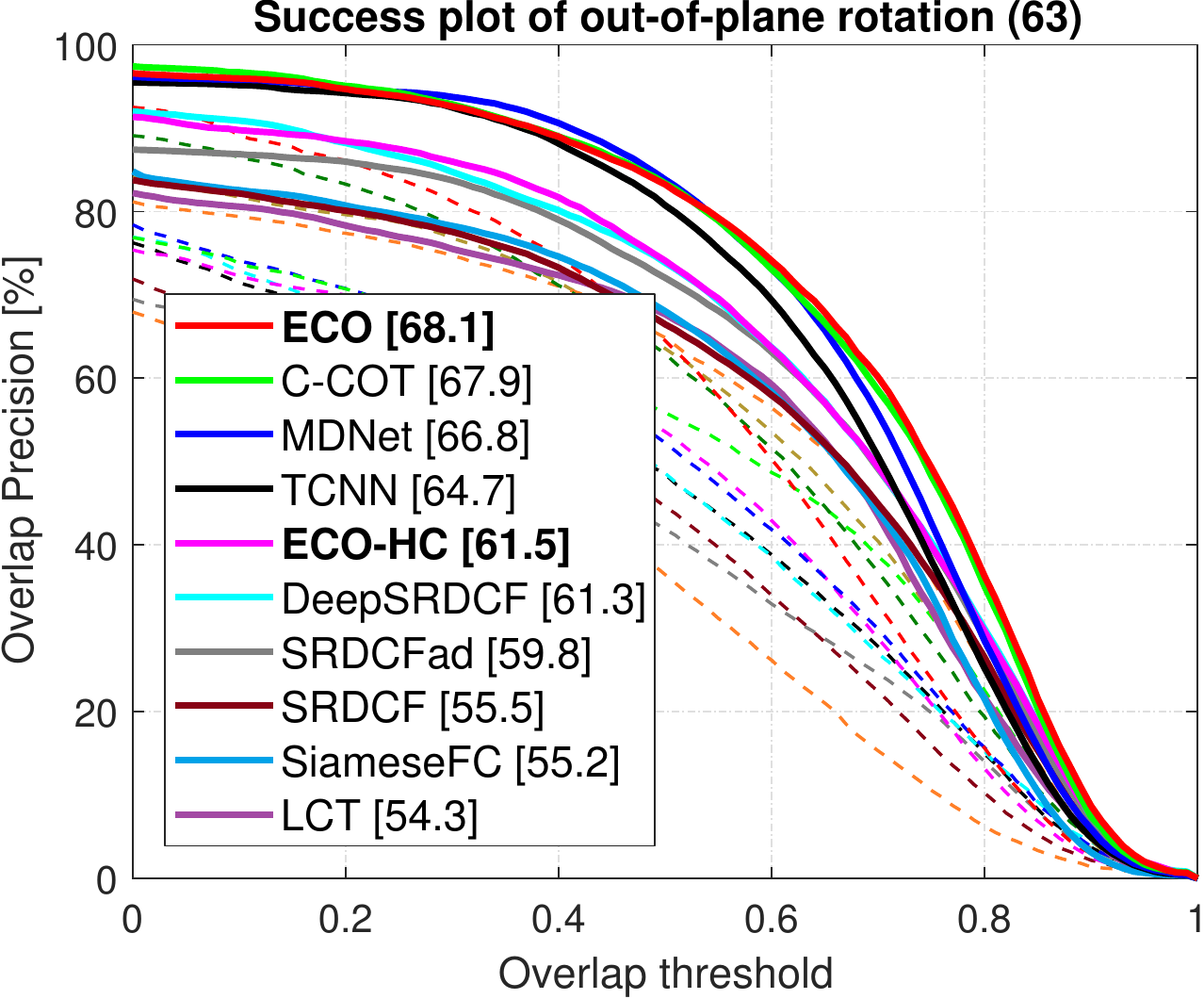}%
	\includegraphics[width=\wid]{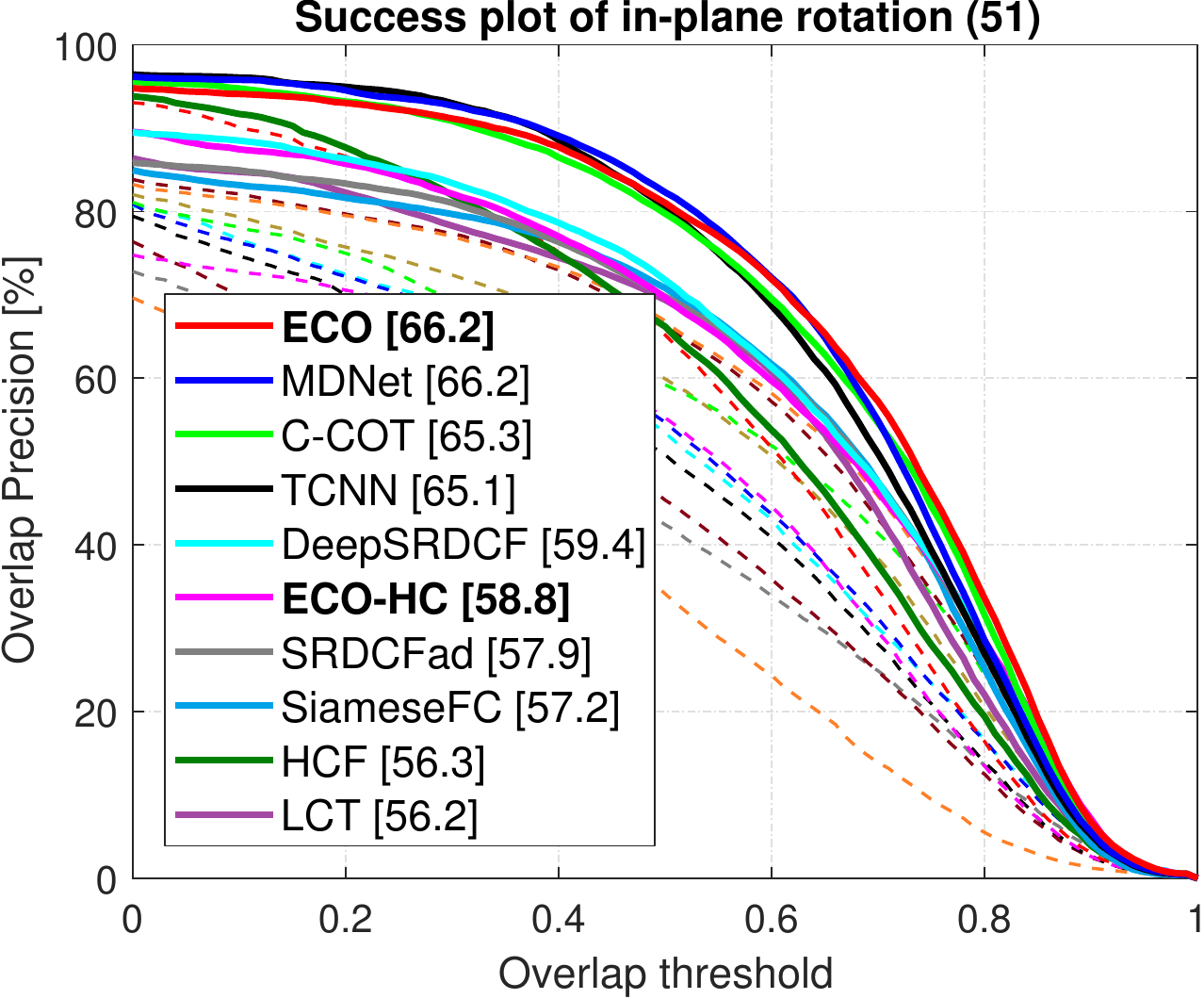}%
	\includegraphics[width=\wid]{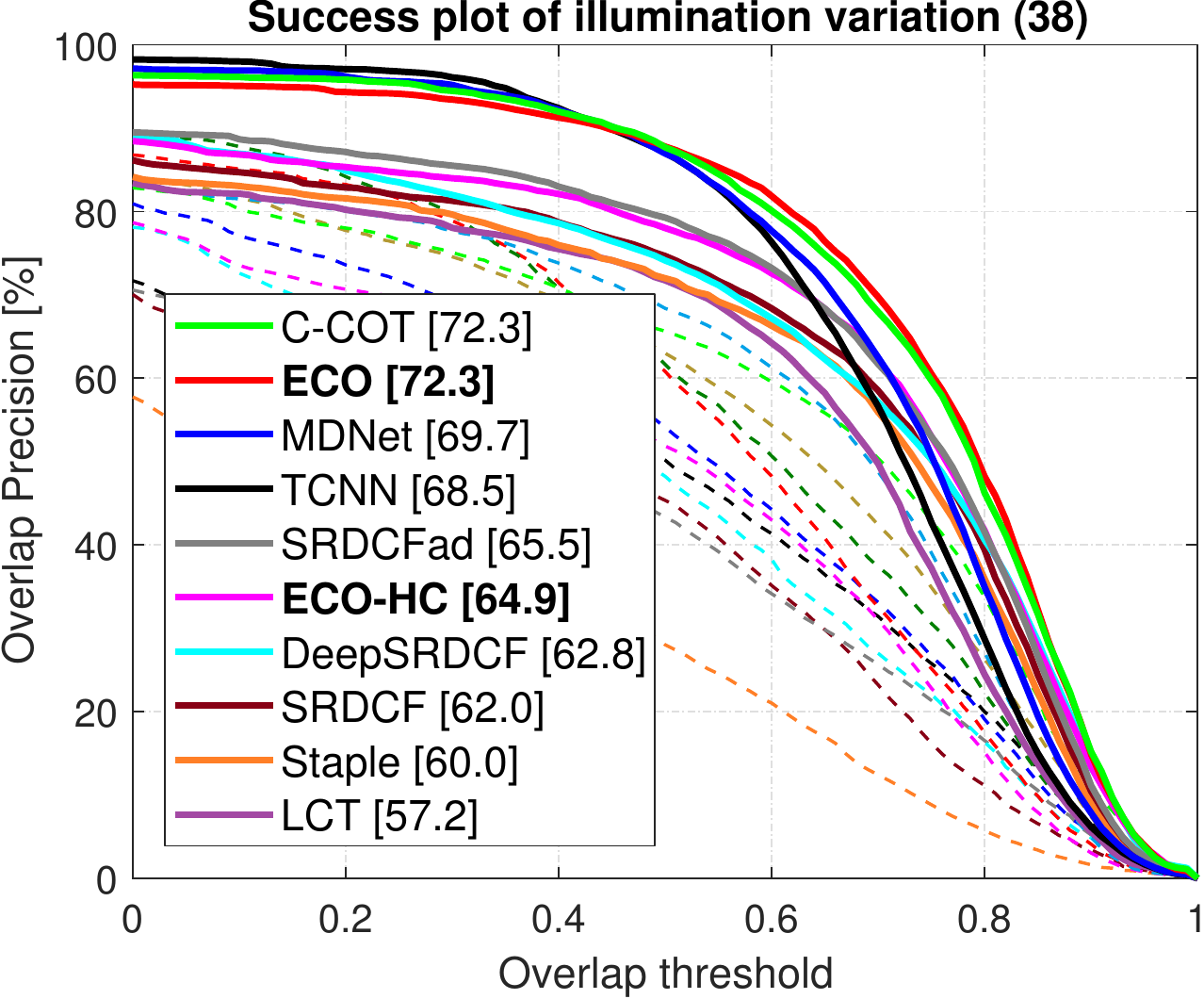}
	\includegraphics[width=\wid]{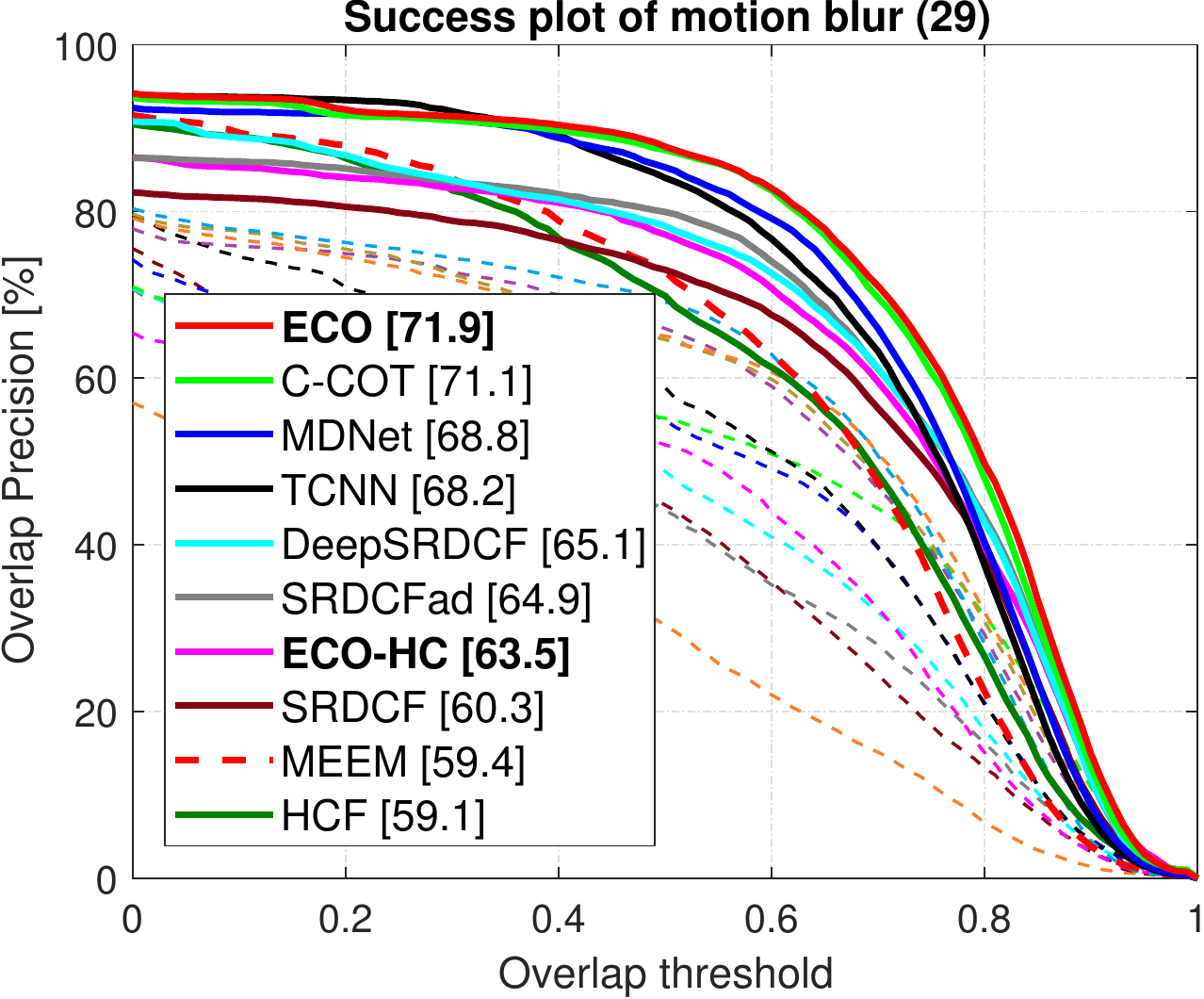}%
	\includegraphics[width=\wid]{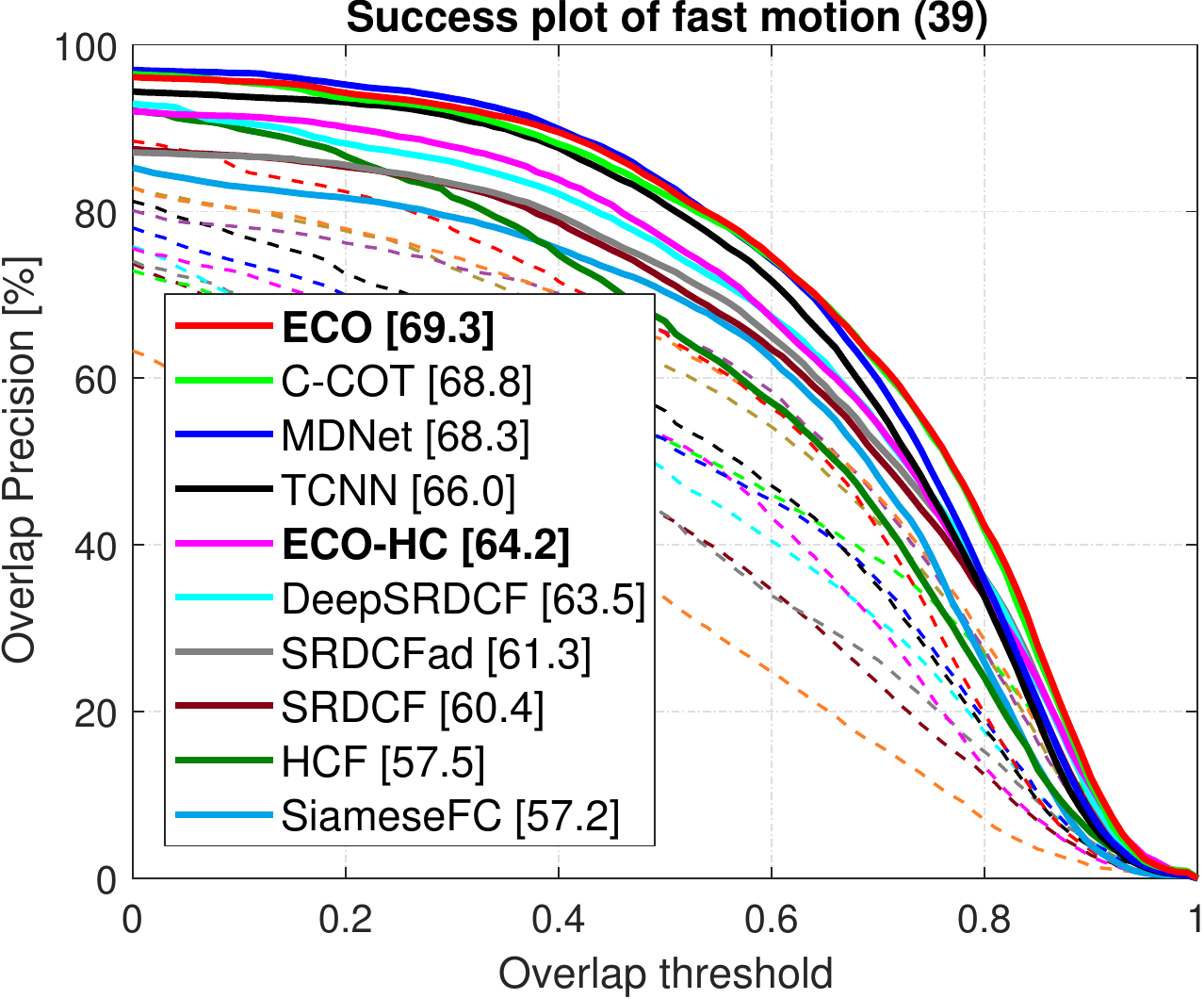}%
	\includegraphics[width=\wid]{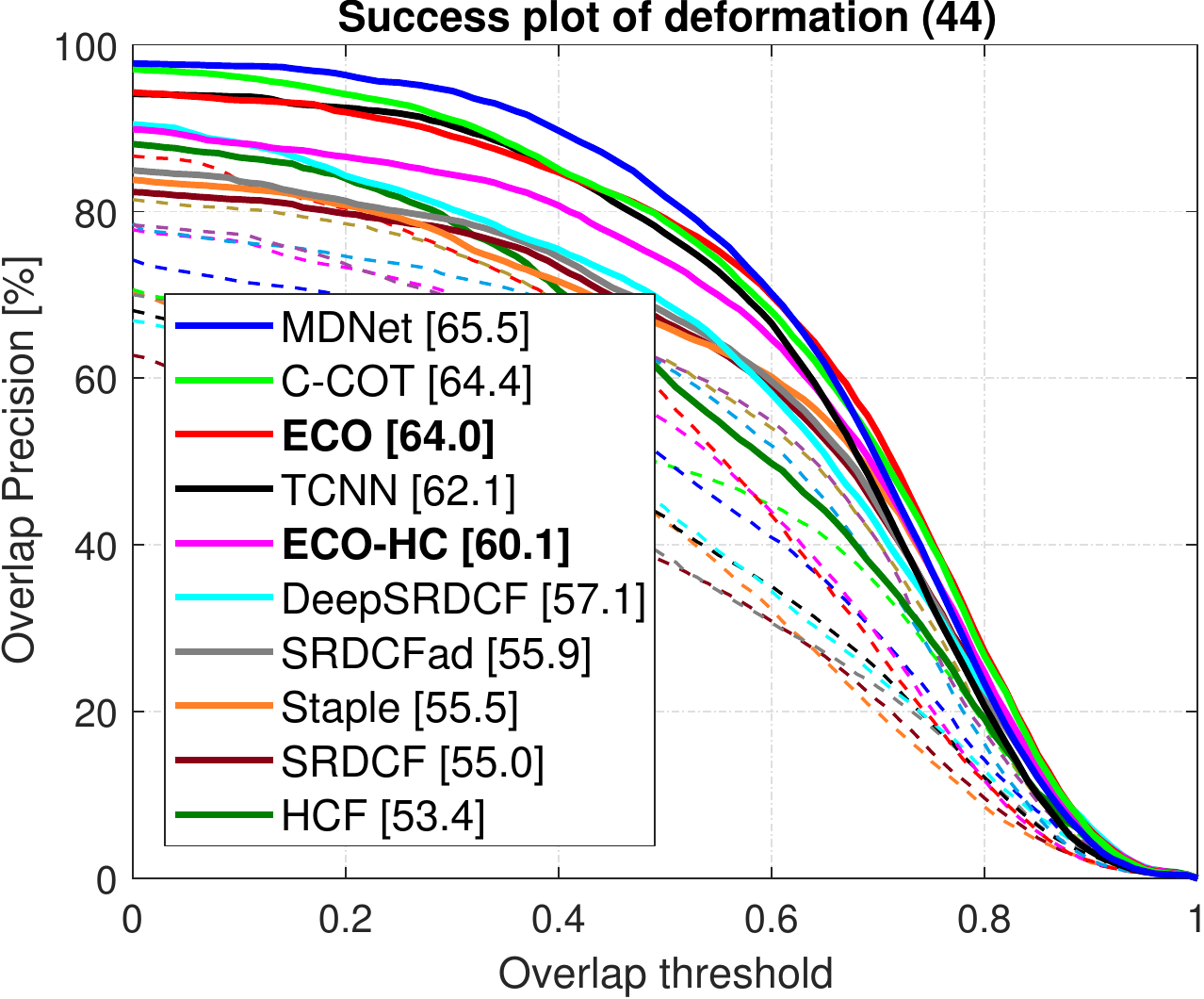}
	\includegraphics[width=\wid]{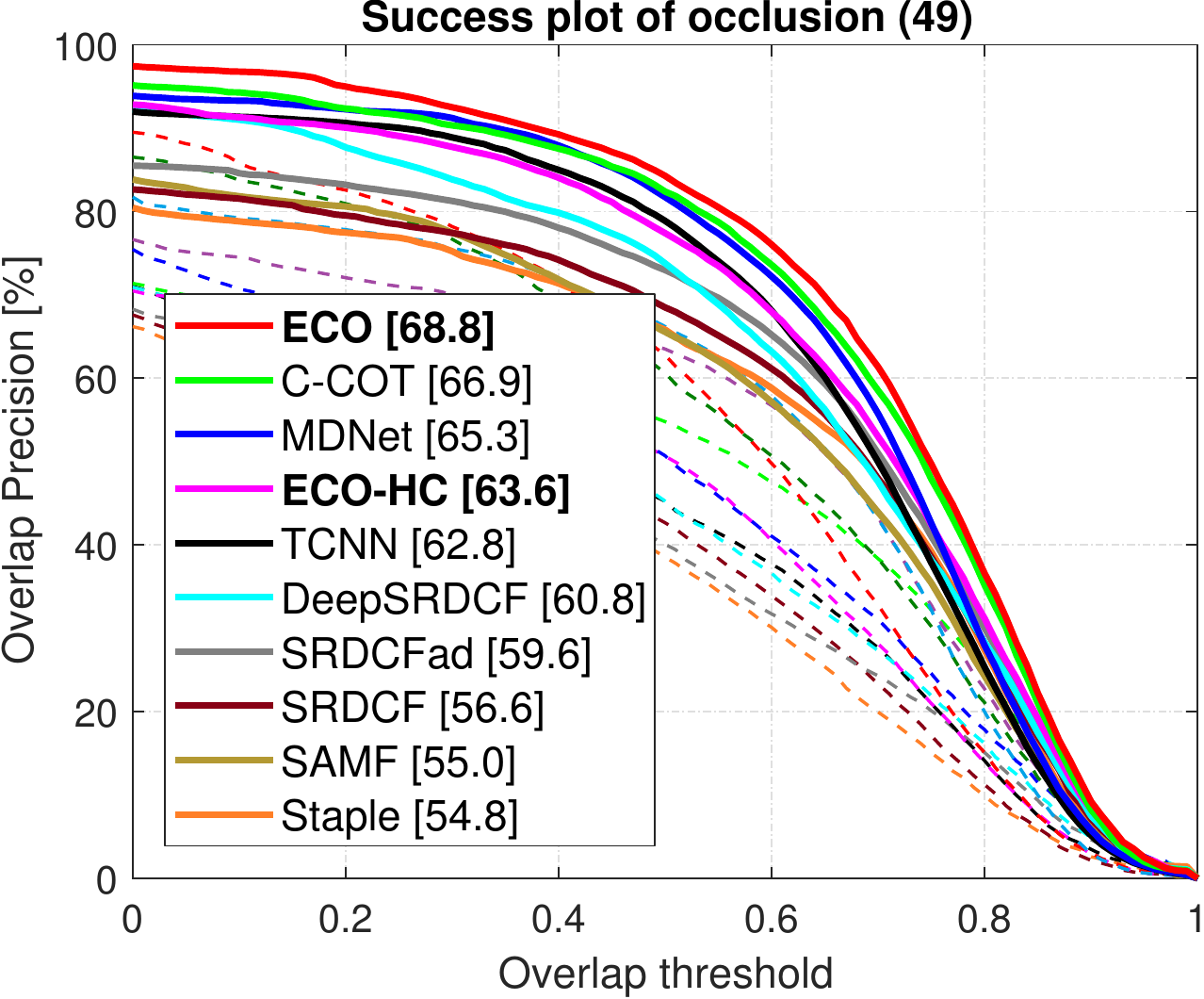}%
	\includegraphics[width=\wid]{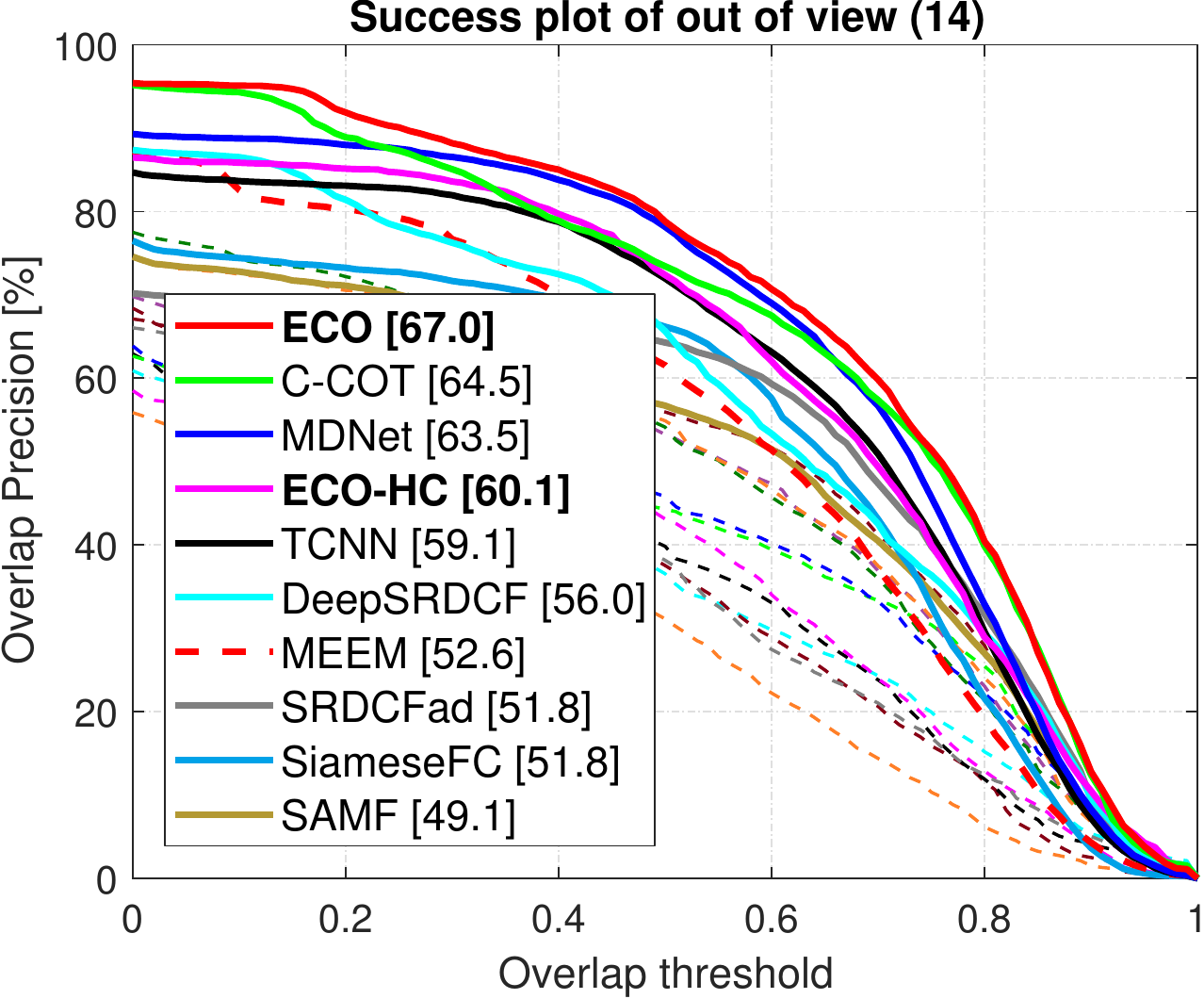}%
	\includegraphics[width=\wid]{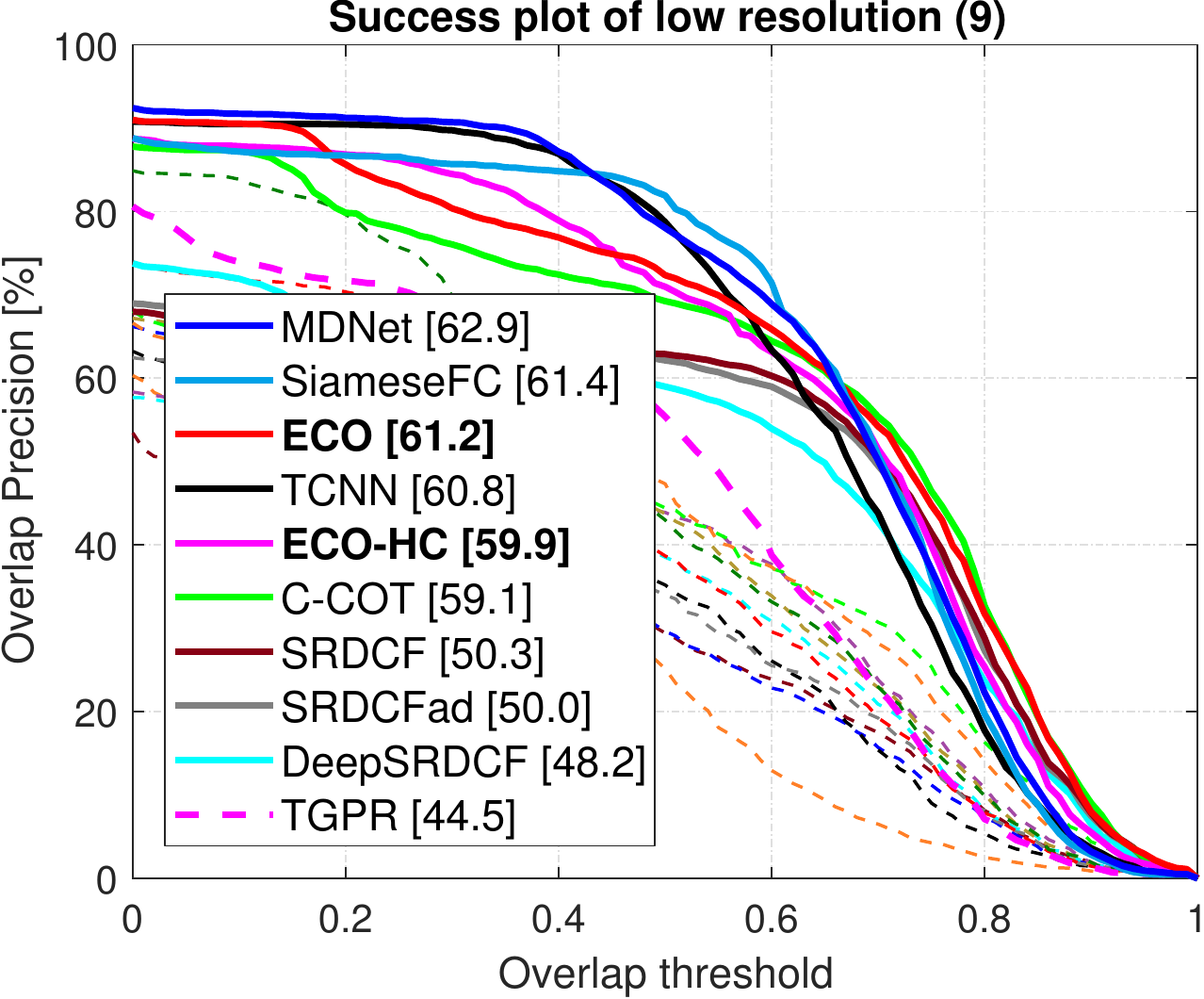}
	\caption{Success plots on the OTB-2015 dataset \cite{OTB2015}. The total success plot (top-left) is displayed along with the plots for all 11 attributes. The title text indicate the name of the attribute and the number of videos associated with it. The area-under-the-curve scores for the top 10 trackers are shown in the legend.}
	\label{fig:attribute}
\end{figure*}

In the OTB-2015 dataset, each video is annotated with 11 different attributes: scale variation, background clutter, out-of-plane rotation, in-plane rotation, illumination variation, motion blur, fast motion, deformation, occlusion, out of view and low resolution. The success plots of all attributes are shown in figure~\ref{fig:attribute}. Our ECO tracker achieves the best performance on 8 out of 11 attributes. Further, our method improves over the baseline C-COT \cite{DanelljanECCV2016} on 9 out of 11 attributes. For a fair comparison, we employ the same combination of deep and hand-crafted features in the baseline C-COT and as in our ECO tracker on the OTB, TempleColor and UAV123 datasets (Conv1, Conv5 and HOG). Note that this set of features provides substantially improved performance in C-COT compared to the original results reported in \cite{DanelljanECCV2016}, where only deep convolutional features are used.

\end{document}